\documentclass[10pt,twocolumn,letterpaper]{article}

\usepackage{iccv}
\usepackage{times}
\usepackage{epsfig}
\usepackage{graphicx}
\usepackage{amsmath}
\usepackage{amssymb}
\usepackage{booktabs}
\usepackage{color,colortbl}
\usepackage{pifont}
\usepackage{subcaption}
\usepackage{tabularx}
\usepackage{multirow}
\usepackage{soul}
\usepackage{enumitem}
\usepackage[accsupp]{axessibility} % Improves PDF readability for those with disabilities.
\usepackage[title]{appendix}

\usepackage[breaklinks=true,bookmarks=false]{hyperref}

\iccvfinalcopy % *** Uncomment this line for the final submission

\usepackage[capitalize]{cleveref}
\crefname{section}{Sec.}{Secs.}
\Crefname{section}{Section}{Sections}
\Crefname{table}{Table}{Tables}
\crefname{table}{Tab.}{Tabs.}

\definecolor{demphcolor}{RGB}{100,100,100}
\definecolor{citecolor}{RGB}{0,0,192} % \definecolor{citecolor}{RGB}{34,139,34}
\definecolor{GrayBG}{gray}{0.95}

\newcommand{\cmark}{\ding{51}}%
\newcommand{\xmark}{\ding{55}}%

\newcommand{\app}{\raise.17ex\hbox{$\scriptstyle\sim$}}
\newlength\savewidth\newcommand\shline{\noalign{\global\savewidth\arrayrulewidth
  \global\arrayrulewidth 1pt}\hline\noalign{\global\arrayrulewidth\savewidth}}

\def\iccvPaperID{2298} % *** Enter the ICCV Paper ID here

\ificcvfinal\fi

\newcommand{\tadet}[0]{TA-IDet}
\newcommand{\tadetspace}[0]{TA-IDet }

\begin{document}

%%%%%%%%% TITLE
\title{EgoObjects: A Large-Scale Egocentric Dataset for Fine-Grained \\Object Understanding}

\author{Chenchen Zhu, Fanyi Xiao, Andrés Alvarado, Yasmine Babaei, Jiabo Hu \\ Hichem El-Mohri, Sean Chang Culatana, Roshan Sumbaly, Zhicheng Yan \\
Meta AI \\
{\tt\small \{chenchenz, fanyix, zyan3\}@meta.com}\\[0.2cm]
\href{https://github.com/facebookresearch/EgoObjects}{\small{https://github.com/facebookresearch/EgoObjects}}
}
% \and

\twocolumn[{
\maketitle
\ificcvfinal\thispagestyle{empty}\fi

\vspace{-0.8cm}
\begin{center}
    % \centering
    \includegraphics[width=2.05\columnwidth]{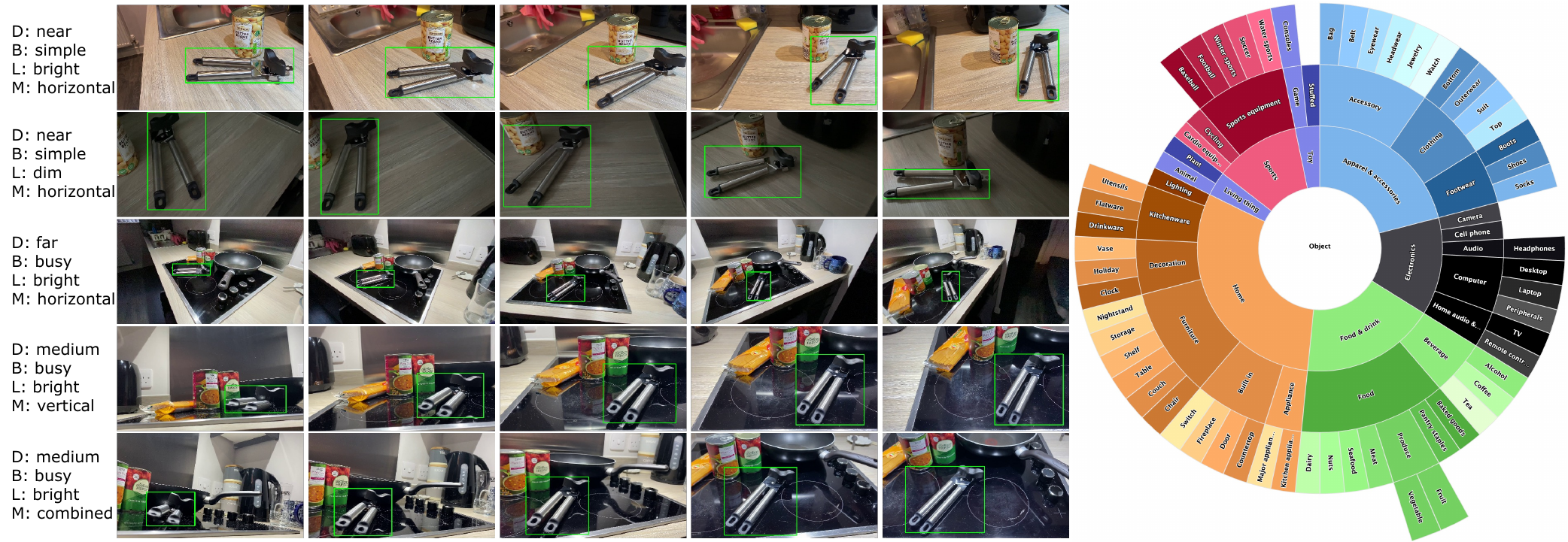}
    \captionof{figure}{ \it \textbf{EgoObjects dataset}. \textbf{Left}: It contains videos of objects captured from the first-person viewpoint under 10 diverse conditions (only 5 are shown for clarity). Multiple objects in each video are annotated with instance ID and category label. In each row, we visualize the annotations of one instance track (``can opener'') in one video captured under one set of condition variable choices. 
    For clarity, we use shorthand notations:  D -- Distance, B -- Background, L -- Lighting, M -- Camera Motion.  
    Also annotations on other objects are not shown. 
    \textbf{Right:} A visualization of a subset of non-leaf nodes in our hierarchical object taxonomy, covering diverse object categories. Leaf nodes and other non-leaf nodes are omitted for clarity. 
    }
    \label{fig:data_intro}
\end{center}

}]
%%%%%%%%% ABSTRACT

\begin{abstract}

Object understanding in egocentric visual data is arguably a fundamental research topic in egocentric vision. However, existing object datasets are either non-egocentric or have limitations in object categories, visual content, and annotation granularities. In this work, we introduce EgoObjects, a large-scale egocentric dataset for fine-grained object understanding. Its Pilot version contains over 9K videos collected by 250 participants from 50+ countries using 4 wearable devices, and over 650K object annotations from 368 object categories. Unlike prior datasets containing only object category labels, EgoObjects also annotates each object with an instance-level identifier, and includes over 14K unique object instances. EgoObjects was designed to capture the same object  under diverse background complexities, surrounding objects, distance, lighting and camera motion. In parallel to the data collection, we conducted data annotation by developing a multi-stage federated annotation process to accommodate the growing nature of the dataset. To bootstrap the research on EgoObjects, we present a suite of 4 benchmark tasks around the egocentric object understanding, including a novel instance level- and the classical category level object detection. Moreover, we also introduce 2 novel continual learning object detection tasks. The dataset and API are available at \href{https://github.com/facebookresearch/EgoObjects}{https://github.com/facebookresearch/EgoObjects}.

\end{abstract}

\vspace{-0.5cm}
\section{Introduction}
\label{sec:intro}
Object understanding tasks, such as classification and detection, are arguably  fundamental research topics in computer vision. Enormous amount of advances achieved so far have been accelerated by the availability of large-scale datasets, such as ImageNet~\cite{imagenet}, COCO~\cite{coco}, LVIS~\cite{lvis}, Open Images~\cite{openimages} and  Objectron~\cite{objectron}. Those datasets often contain images captured from a third-person or exocentric viewpoint and curated from given sources (e.g. Flicker). 
Albeit the large volume, they often only capture individual object instances in a single image or video, and do not capture the same object  under diverse settings, which are important for fine-grained object understanding task, such as instance-level object detection. In contrast, object understanding tasks in egocentric vision processes visual data containing objects captured from a first-person or egocentric viewpoint. The approaches to those tasks have wide applications in augmented reality and robotics, such as robustly anchoring virtual content at a real world object under various conditions (e.g. background, lighting, distance), and are often required to perform well from the egocentric viewpoint and distinguish objects at both category- (e.g. mug vs kettle) and instance level (e.g. my mug vs your mug) under various conditions. Therefore, there are clear gaps in adopting existing exocentric datasets for egocentric object understanding.

On the other hand, several egocentric datasets containing object annotations have been built. A family of such datasets are focused on capturing human activities and hand-object interactions. Ego4D~\cite{ego4d} contains a large number of egocentric videos of human activities. However, according to the PACO-Ego4D \cite{paco} which mines the objects from Ego4D, there are only 75 object categories with at least 20 samples, and each object instance often only appears in one video. Epic-Kitchens-100~\cite{epic-kitchens-100} contains over 700 videos depicting human activities in the kitchen, but only annotates objects within the kitchen. 
HOI4D~\cite{hoi4d} is collected for category-level human-object interaction, and only contains 800 different object instances from 16 categories.
There are several other datasets that are more object-centric, including TREK-150~\cite{trek150}, FPHA~\cite{fpha} and CO3D~\cite{co3d}, but only contain objects from a limited set of categories ($<$50). Objects there are often captured in a single setup or few setups with limited variations in surrounding objects, background, distances and camera motions. Moreover, semantic granularity of the object annotations are often limited at category-level, and object instances from the same category are not distinguished, which impedes the development of instance-level object understanding approaches. Therefore, there are still significant gaps with existing egocentric datasets in the dataset scale, visual content variations around individual objects, object semantic diversity, and instance-level object annotation.

To address these gaps, we introduce \textit{EgoObjects}, a new large-scale egocentric video dataset for fine-grained object understanding (Figure \ref{fig:data_intro}). Unlike prior egocentric datasets which are limited to a small dataset scale, a specific domain or a small number of object categories, EgoObjects includes a large number of videos containing objects from hundreds of object categories commonly seen in the households and offices worldwide. For video capture, 4 wearable devices with various field-of-view are used, including Vuzix Blade smart glasses\footnote{\href{https://www.vuzix.com}{https://www.vuzix.com}  }, Aria glasses\footnote{\href{https://about.meta.com/realitylabs/projectaria}{https://about.meta.com/realitylabs/projectaria} }, Ray-Ban Stories smart glasses\footnote{\href{https://www.meta.com/glasses}{https://www.meta.com/glasses}} and mobile phones with ultra-wide lens\footnote{Participants are asked to hold mobile phone close to their eyes to simulate egocentric viewpoints}, which provide representative media formats of egocentric visual data. Each main object is captured in multiple videos with different choices of nearby secondary objects, background complexity, lighting, viewing distance and camera motion. We annotate both the main and secondary objects in the sampled frames with bounding boxes, category level semantic labels and instance-level object identifiers (ID). In current Pilot version release, it contains over $9,200$ videos of over 30 hours collected by 250 participants from 50+ countries and regions, and 654K object annotations with 368 object categories and 14K unique object instance IDs from 3.8K hours of annotator efforts. To our best knowledge, EgoObjects is the largest egocentric video dataset of objects in terms of object categories, videos with object annotations, and object instances captured in multiple conditions. 
Comparisons between EgoObjects and other datasets can be seen in Table~\ref{tab:comparison}.

To bootstrap the %object understanding%
research on EgoObjects, we introduce 4 benchmark tasks spanning over both non-continual learning and continual learning settings. For non-continual learning setting, we include a novel instance-level object detection task, largely under-explored previously due to the lack of a dataset with object ID annotations, as well as conventional category-level object detection task. For continual learning setting, we present novel object detection tasks at instance- and category level. Evaluations of different approaches to all tasks are also presented to establish the baseline benchmarks. In particular, for instance-level object detection task, a novel target-aware instance detection approach is proposed and validated to outperform a baseline target-agnostic object detection method.

To summarize, we make the following contributions.

\begin{itemize}[leftmargin=*]

\item We created a large-scale egocentric dataset for object understanding, which features videos captured by various wearable devices  at worldwide locations, objects from a diverse set of categories commonly seen in indoor environments, and videos of the same object instance captured under diverse conditions.

\item We proposed a multi-stage federated annotation process for the continuously growing dataset to accompany the parallel data collection at scale. Rich annotations at video level (e.g. location, background description) and object-level (e.g. bounding box, object instance ID, category level semantic label) are collected from 3.8K hours of human annotator efforts.

\item We introduced 4 benchmark tasks on EgoObjects, including the novel instance-level and the conventional category-level object detection tasks  as well as their continual learning variants. We evaluated multiple approaches on all tasks, and also proposed a novel target-aware approach for instance-level object detection task.

\end{itemize}

\begin{table}

\centering

\setlength{\tabcolsep}{2pt}
\resizebox{1.02\columnwidth}{!}{
\begin{tabular}{l|lll|llll}
 & \multicolumn{3}{c|}{Exocentric} & \multicolumn{4}{c}{Egocentric} \\ %\cline{2-8} 
 & Objectron & CO3D & BOP & Epic-K$^{*}$. & HOI4D & Ego4D$^{**}$ & EgoObjects \\ \shline
\#category & 9 & 50 & 89 & 300$^{*}$ & 16 & 75 & 368+ \\
\#participant & int'l. & - & - & 45 & 9 & 859 int'l. & 250 int'l. \\
\#image & 4M & 1.5M & 330K & 20M & 2.4M & 23.9K & 114K+ \\
\#instance & 17K & 19K & 89 & - & 800 & 17K & 14K+ \\
\#bbox & - & - & - & 38M & - & 50K & 654K+ \\
inst ID & \xmark & \xmark & \cmark & \xmark & \xmark & \xmark & \cmark \\

device & M & M & PC,K & G & K,I & G,V,Z,W,PP & R,A,V,M \\ 
\hline
\end{tabular}
}
% \vspace{-0.2cm}
\caption{
\small \it \textbf{Comparing EgoObjects with other datasets.}
For EgoObjects, we report statistics of the current Pilot version, which is estimated to account for 10\% of the full dataset (thus the ``+'' notation).
$^*$Epic-Kitchen-100~\cite{epic-kitchens-100} only contain object categories in the kitchen.
$^{**}$Ego4D statistics are reported by the PACO-Ego4D~\cite{paco}, which annotates objects in the Ego4D~\cite{ego4d}.
Abbreviation for devices: M=Mobile, K=Kinect, A=Aria, G=GoPro, PC=Primesense Carmine, I=Intel RealSense, V=Vuzix Blade, R=Ray-Ban Stories, PP=Pupil, Z=Zetronix zShades, W=Weeview. 
}
\label{tab:comparison}
% \vspace{-0.5cm}
\end{table}

\section{Related Work}
\label{sec:related}

\begin{figure*}
% \vspace{-0.8cm}

    \centering
    \begin{subfigure}[t]{0.7\columnwidth}
        \centering
        \includegraphics[width=\textwidth]{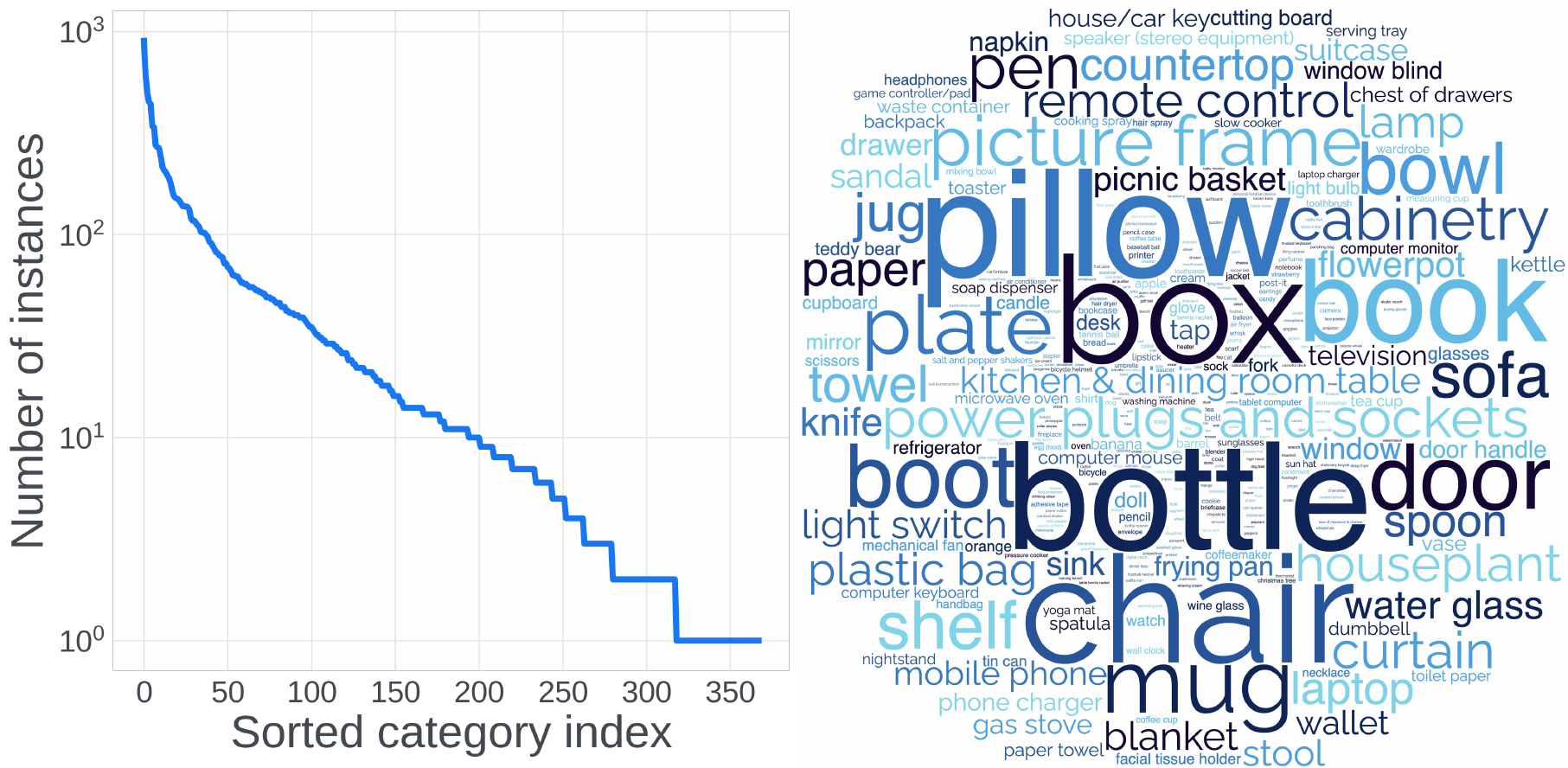}

        \caption{\small \it Number of instances per category.}
        \label{fig:dist_inst}
    \end{subfigure}
    \quad % \hspace{0mm}
    \begin{subfigure}[t]{0.7\columnwidth}
        \centering
        \includegraphics[width=\textwidth]{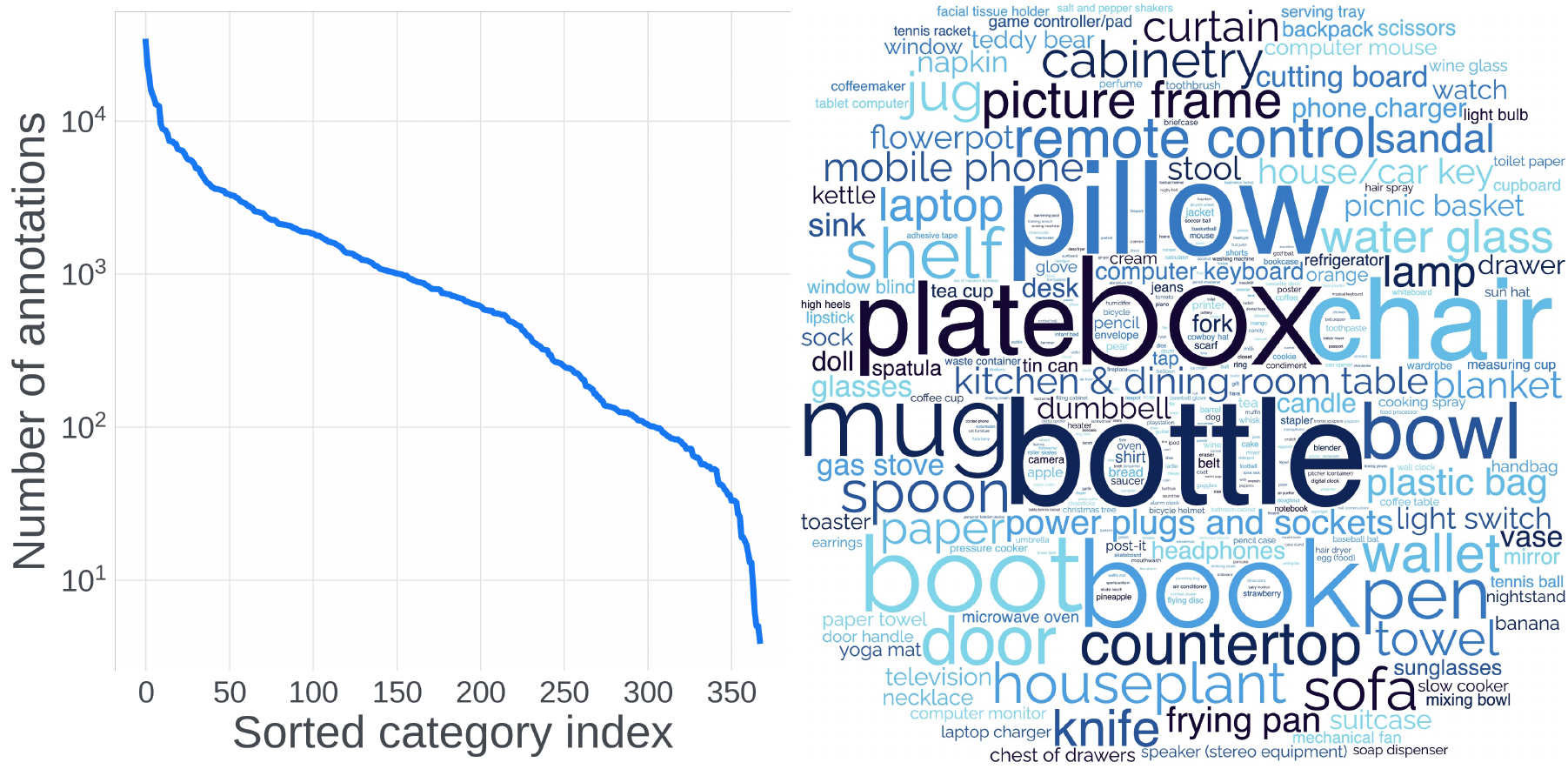}
 
        \caption{\small \it Number of annotations per category.}
        \label{fig:dist_anno}
    \end{subfigure}
    \quad % \hspace{0mm}
    \begin{subfigure}[t]{0.5\columnwidth}
        \centering
        \includegraphics[width=0.7\textwidth]{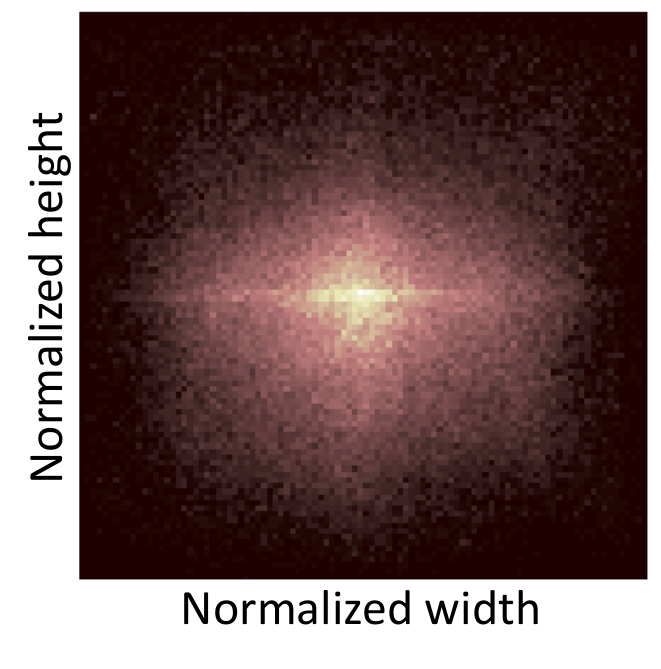}

        \caption{\small \it Distribution of object centers.}
        \label{fig:spatial_stats}
    \end{subfigure}
    
    \begin{subfigure}[t]{0.65\columnwidth}
        \centering
        \includegraphics[width=0.9\textwidth]{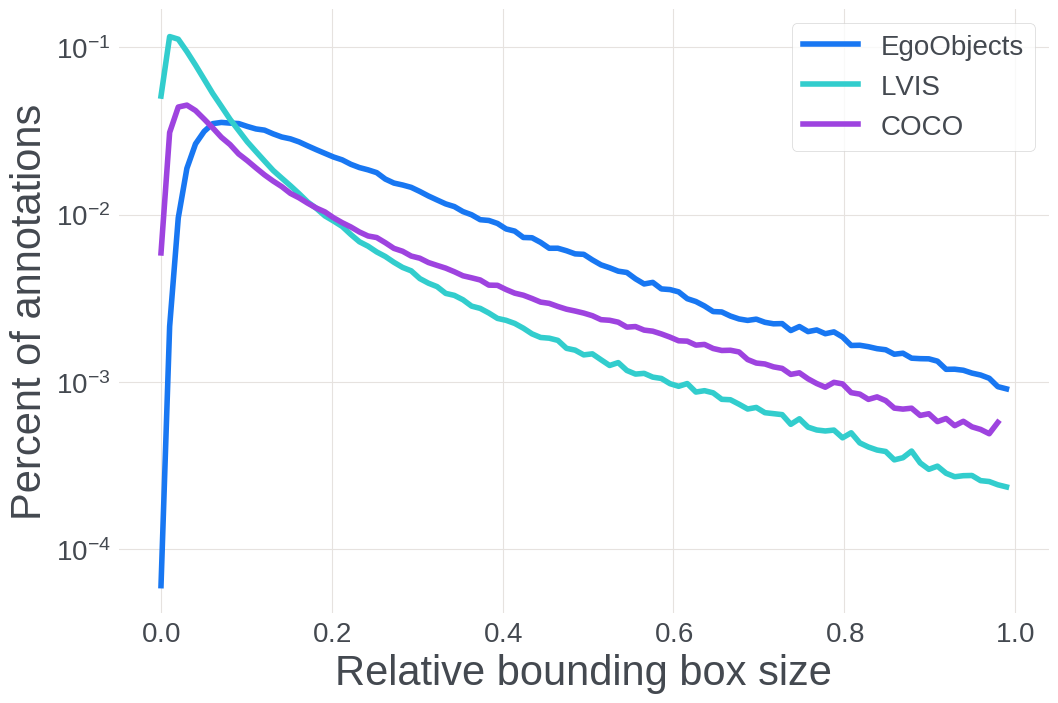}
        \caption{\small \it Relative bounding box sizes.}
        \label{fig:scale_stats}
    \end{subfigure}
    \quad
    \begin{subfigure}[t]{0.6\columnwidth}
        \includegraphics[width=\textwidth]{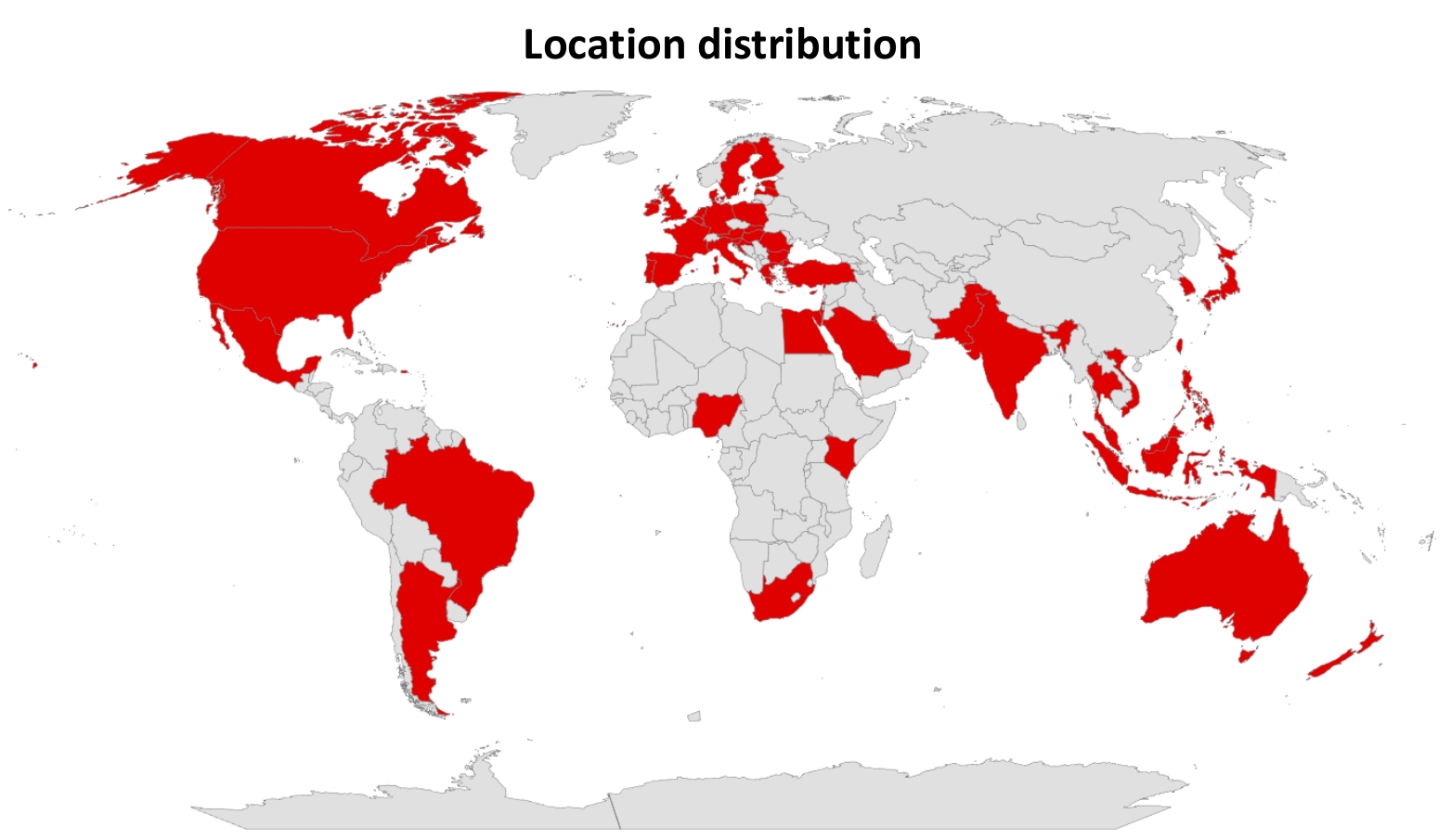}
        \caption{\small \it Geo distribution of participants.}
        \label{fig:dist_location}
    \end{subfigure}
    \quad
    \begin{subfigure}[t]{0.65\columnwidth}
        \includegraphics[width=\textwidth]{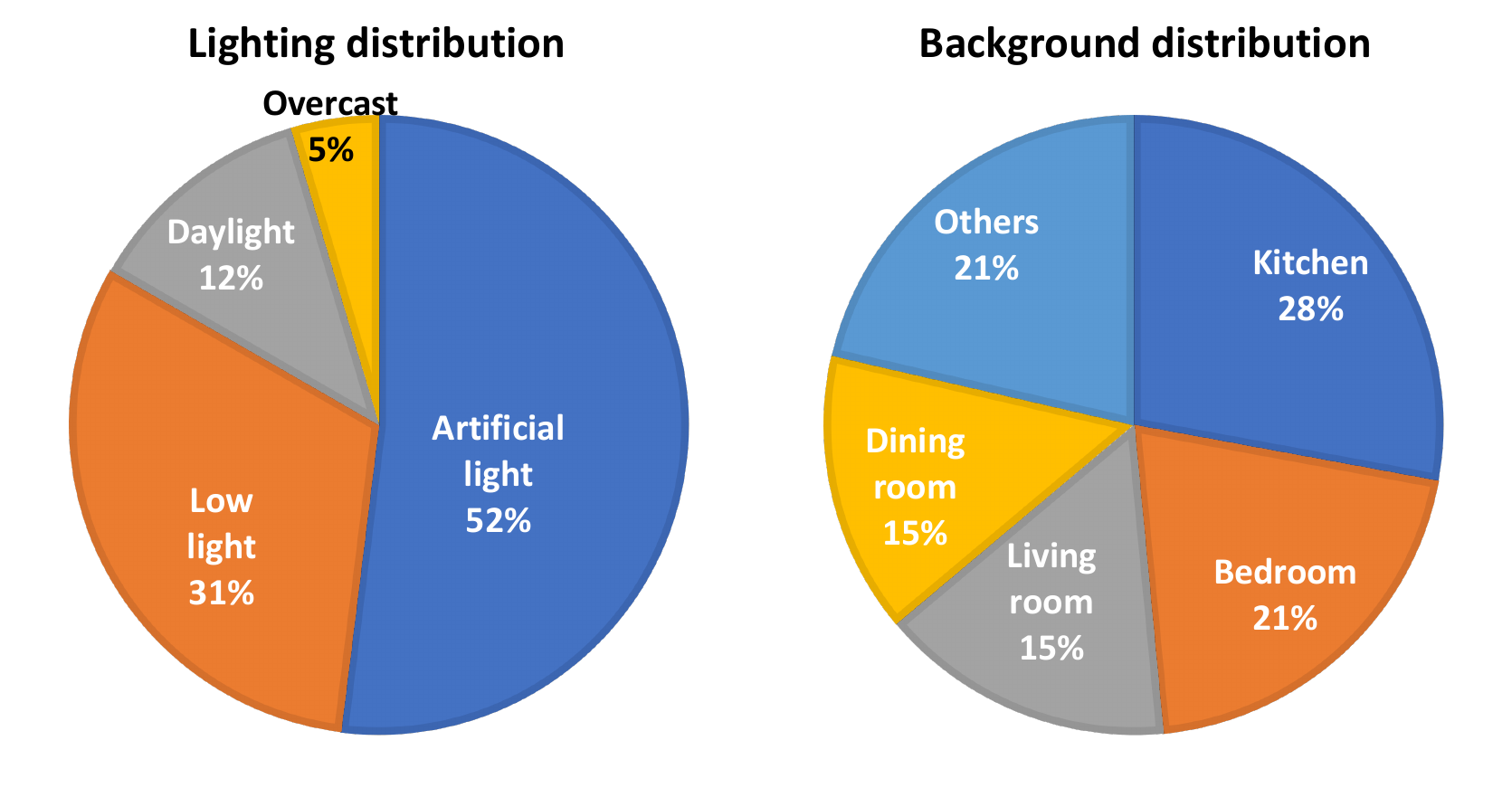}
        \caption{\small \it Distribution of video metadata.}
        \label{fig:metadata_stats}
    \end{subfigure}
    % \vspace{-0.3cm}
    \caption{\small \it \textbf{Dataset statistics.} 
    \textbf{(a)} \textbf{Left}: the number of instances per category in the log scale. \textbf{Right}: the word cloud highlights the head categories, including box, bottle.
    \textbf{(b)} \textbf{Left} the number of annotations per category in the log scale. \textbf{Right}: the word cloud is similar to (a), but a few new head categories emerge, including mug, boot.
    \textbf{(c)} Spatial distribution of the main objects center coordinates, confirming the diverse locations of main objects.
    \textbf{(d)} Relative bounding box sizes compared between EgoObjects, LVIS, and COCO. EgoObjects has more objects of medium and large sizes in the egocentric view.
    \textbf{(e)} Diverse distribution of participants' geographic locations in 50+ countries from 5 continents.
    \textbf{(f)} Distribution of video metadata including lighting (left) and background (right). Most objects are collected indoor, where lighting is more likely either artificial or low light. The background is uniformly distributed across rooms within the household.
    % Best viewed in pdf, zoomed in. 
    }
    \label{fig:inst_stats}

% \vspace{-0.5cm}
\end{figure*}

\noindent \textbf{Egocentric object understanding datasets.}
Given the growing needs of egocentric object understanding in augmented reality and robotics, several egocentric datasets focused on objects have been built. TEgO \cite{tego} contains egocentric images of only 19 distinct objects for training object recognizers. TREK-150 \cite{trek150} consists of 150 annotated videos for tracking objects from 34 categories merely. 
Despite the availability of object annotations, other larger egocentric datasets are more focused on human activities and hand-object interactions. For example, Epic-Kitchens-100~\cite{epic-kitchens-100} captures 700 videos of nearly 100 human activities involving 300 object categories in the kitchen, but is limited to the kitchen scenario. The ADL \cite{adl} dataset features people performing everyday activities in kitchens, which has object boxes, object track ID, action labels. However, it only has 42 object categories and the track ID is not used for analysis. The MECCANO \cite{MECCANO} is a multimodal dataset of egocentric videos to study humans behavior understanding in industrial-like settings with object, depth, and gaze annotations, supporting a suite of 5 tasks. However, the diversity of participants and locations is limited.
FPHA~\cite{fpha} captures 45 different daily hand-object action categories involving only 26 different objects. HOI4D~\cite{hoi4d} contains videos of human-object interaction with only 800 different object instances from 16 categories. 
Albeit the large number of human activity videos, the recent Ego4D~\cite{ego4d} only contains object annotations from around 75 object categories with at least 20 samples. Object instances often only appear in a single video, and only 870 instances have more than 5 occurrences. Meanwhile, synthetic egocentric datasets are built to scale up the data collection. xR-EgoPose \cite{xregopose} is a large-scale synthetic dataset containing realistic renderings of people in various poses and serves as a benchmark of 3D human pose estimation. It is focused on ego-body and simulates fisheye lens where the surrounding environment, including objects, are largely distorted. EHOI \cite{ego-hoi-synthetic} is also a synthetic dataset, consisting of 20K images and 124K object instances from 19 categories with interactions with human hands. Its fidelity is low compared with real data, and it has limited complexities in lighting, background and viewpoints. To summarize, existing egocentric datasets have limitations in the number of object categories, the variations in the setting of capturing the same object, the granularity of object semantic labeling where instance-level object ID is not available and photorealism in synthetic datasets.

\noindent \textbf{Instance-level object detection and datasets.} Being able to localize and recognize different object instances
%, regardless of the object category, 
is critical to applications in augmented reality and robotics, such as detecting a specific toy or a custom industrial part. However, such task has been severely less explored due to the lack of object ID annotations at scale in existing datasets. In the cases of a growing number of object instances to detect, which is arguably a realistic setup, instance-level detection approaches are often required to adapt with little-to-no fine-tuning time. Mercier et al \cite{mercier2021} proposed a template-based detector that uses example viewpoints of the target object to detect it in query images without extra training, and evaluated it on a small exocentric dataset of 20 object instances only. Hu et al~\cite{hu2022template} proposed a template-based detection approach, which incorporated a multi-level correlation model and a similarity-refine module, for handling the category-agnostic instance. On the dataset side, T-less~\cite{tless} is an object dataset with 6D pose annotation for only 30 industry-relevant object instances. In ~\cite{rennie2016dataset}, a small dataset of 10K RGBD images of 24 object instances were created for object detection and pose estimation. BOP dataset~\cite{bop} combines 8 public datasets, and consists of 89 object instances with 3D groundtruth and over 330K RGBD images from different viewpoints. All those datasets aforementioned are not egocentric, and only contains a small number of object instances. In contrast, EgoObjects contains over 14K unique object instances captured under diverse settings. We also propose a target-agnosic baseline approach and a novel target-aware approach, and evaluate them on EgoObjects.

\noindent \textbf{Continual learning.} Conventional object understanding approaches build static models incapable of adapting their predictive behaviors over time. In contrast, continual learning models can learn from an infinite stream of data and grow their predicative capabilities while reducing catastrophic forgetting of previous knowledge~\cite{masana2022class, buzzega2020dark, bang2021rainbow, chen2018lifelong, parisi2019continual, aljundi2018selfless, rebuffi2017icarl, castro2018end, van2019three}. Broadly speaking, they can be categorized into 3 classes~\cite{CL-survey} with increasing complexities and practicalities. In \textit{Task Incremental Learning}, individual tasks with respective training data arrive sequentially, and the model is often built with separate heads for individual tasks. At inference time, a task ID is required for each sample. In the \textit{Class Incremental Learning}, no task ID is provided at any time, and the model often has only one head. In the most general \textit{Data Incremental Learning}~\cite{de2021continual}, more assumptions on the stationary data distribution and the paradigm of sequentially growing tasks and classes are removed. In this work, we use EgoObjects to set up 2 new continual learning tasks, which covers both Class- and Data Incremental Learning paradigms. Moreover, existing approaches are often assessed on small object classification datasets, such as Core50~\cite{core50} and OpenLORIS-Object~\cite{Openloris-object}. To our best knowledge, EgoObjects is the first dataset to support the benchmarking of \textit{continual learning of object detection} at both instance and category level.

\noindent \textbf{Category-level object detection.}  Early CNN-based approaches include both two-stage methods, which tackles object proposal generation and object recognition separately~\cite{faster-rcnn,cascade-rcnn}, and single-stage methods which remove the explicit object proposal generation for simplicity and efficiency~\cite{liu-ssd2016,Yolov4,tian-fcos2019}. Recent transformer-based methods introduce attention building blocks into both the backbone and the detection head to significantly improve the detection performance~\cite{carion-detr2020,zhu-deformabledetr2020,dai-dynamicdetr2021}. However, those approaches are often only evaluated on exocentric datasets, such as COCO~\cite{coco} and LVIS~\cite{lvis}, while their performance on egocentric datasets are largely unknown. EgoObjects contains nearly 400 object categories, and we assess both CNN and transformer models on it.

\section{EgoObjects Dataset}
\label{sec:dataset}

\subsection{An Overview}
In current Pilot version, EgoObjects contains over 9K videos collected by 250 participants. A total of 114K frames are sampled and annotated.

\noindent \textbf{Object instances captured under diverse conditions.}
A total of 14.4K unique object instances from 368 categories are annotated. Among them, there are 1.3K main object instances from 206 categories and 13.1K secondary object instances (\ie, objects accompanying the main object) from 353 categories. 
On average, each image is annotated with 5.6 instances from 4.8 categories, and each object instance appears in 44.8 images, which ensures diverse viewing directions for the object. 
To further break down, for the main object, each instance appears in 95.9 images, whereas each secondary instance \ie 39.8 images on average. See distributions of unique object instances and object annotations in Figure~\ref{fig:dist_inst} and \ref{fig:dist_anno}, respectively.
Both figures indicate the long-tailed nature of the dataset, making the benchmark more challenging and closer to the real-world distributions.

\noindent \textbf{Diverse object spatial distribution}. 
We encourage participants to avoid center bias during the capture of main objects with the moving camera. In Figure~\ref{fig:spatial_stats}, we confirm both the center coordinates are widely spread in the image.

\noindent \textbf{Object scale distribution in egocentric view}. 
Figure \ref{fig:scale_stats} compares EgoObjects with other datasets on the relative size distribution of object bounding boxes. 
The relative size is defined as the square root of the box-area-over-image-area ratio. 
Compared to COCO and LVIS, EgoObjects has more medium and large-sized objects, and suits the applications of egocentric object understanding where the users are more likely to interact with closer objects.

\noindent \textbf{Metadata statistics}. 
We further accumulate the per-video statistics across several metadata tags.
As shown in Figure~\ref{fig:dist_location}, our data is geographically diverse, covering 50+ countries from five continents. 
Finally, as presented in Figure~\ref{fig:metadata_stats}, our data also has a diverse distribution covering various video capture conditions for lighting and backgrounds.

\begin{figure}
% \vspace{-0.5cm}
    \centering
    \includegraphics[width=\columnwidth]{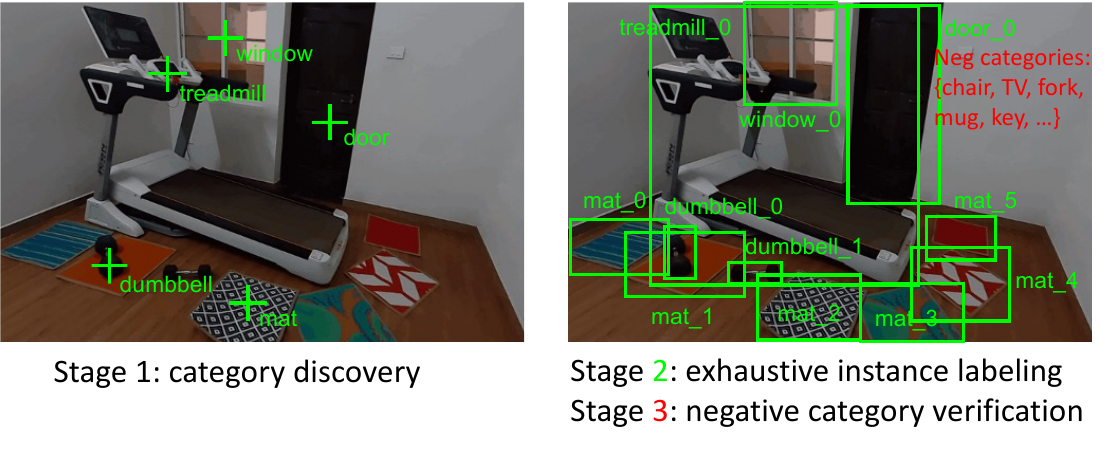}

% \vspace{-0.3cm}

     \caption{\small \it \textbf{EgoObjects multi-stage annotation}. See text for details at each stage.}
    \label{fig:annotation}
% \vspace{-0.5cm}
\end{figure}

\subsection{Data Collection}
\label{sec:dataset:collection}
We work with third-party vendors to recruit participants for capturing videos of common indoor objects at worldwide locations. They use various glasses such as Vuzix Blade, Aria Glasses, and Ray-Ban Stories. They also use the ultra-wide lens on mobile phones and hold the phone close to the eyes for simulating the egocentric viewpoint.
Participants are asked to capture videos of objects from a predefined list of 400 object categories, referred as main object categories. 
Each main object should be unique in the individual location and captured under various conditions. 

We define 4 variables of capture conditions, including background complexity, camera motion, object distance and lighting. The background complexity can be either ``simple'' or ``busy''. The simple background has at least 3 surrounding objects besides the main object, whereas the busy background has at least 5 other objects. In either background, we ask participants to capture the main object in natural settings (vs intentional setting). 
We also instruct the participants to move the camera around and capture different views of the main object, while avoiding the bias that the main object always stays in the center of the view. 
We define three 3 of camera motion: 1) ``horizontal'': move the camera from left to right or right to left. 2) ``vertical'': move the camera upwards or downwards and 3) ``combined'': rotate the camera both horizontally and vertically.
The object distance also has three levels, i.e. ``near'', ``medium'', and ``far''. We define the object scale and the frame scale as the longer object dimension and the shorter frame edge, respectively. 
Near distance refer to those that the object-scale/frame-scale ratio is larger than $30\%$, whereas the medium distance has the ratio fall in between $20\%$ and $30\%$. 
All remaining images are considered as having far object distances to the camera. 
For lighting conditions, there are two levels bright and dim. Lighting is considered as bright when a light meter reads above 250 lux and dim otherwise.

Given these 4 variables, participants are instructed to collect 10 videos of each main object according to 10 predefined configurations (see details in supplement), and each video lasts at least 10 seconds.
Finally, videos are further tagged with rich metadata including the associated participant ID, main object category, location, background description and capture time.

\subsection{Federated Annotation of the Growing Dataset}
\label{sec:dataset:annotation}
EgoObjects data collection was planned to operate at large scale and lasted for 14 months. To reduce the overall dataset creation time, we conducted data annotation in parallel to the data collection, which continuously grew the dataset and introduced more complexities to the data annotation. Inspired by LVIS~\cite{lvis}, we adopt the idea of federated annotation to achieve a balance between annotation cost and annotation exhaustiveness, and further propose a 3-stage annotation pipeline tailored to the continuously growing nature of our dataset. Figure \ref{fig:annotation} illustrates our annotation pipeline, which is used to annotate video frames evenly sampled at 1 FPS. 

\noindent \textbf{Stage 1: category discovery.} 
The annotators are instructed to identify object categories from a predefined vocabulary $\mathcal{V}$ of 600+ categories commonly seen in indoor egocentric view.
Annotators are asked to find at least 5 categories per each image if possible, including the main object category and other salient secondary objects.

\noindent \textbf{Stage 2: exhaustive instance labeling.}
For each image, 3 annotators exhaustively annotate \textit{all} object instances of the discovered categories with bounding box and category label $c$. 
To enable instance-level object understanding, we further enhance the bounding box annotation with a unique object instance ID\footnote{We exclude objects from categories that have indistinguishable appearances between instances, such as those related to animals and food.} that is consistent across the dataset. 
To reconcile the work from 3 annotators, we compare the annotations from one annotator to all other annotators to get an averaged IoU based consensus score for each annotator. Then, we select the annotator with the highest consensus score as the source of truth for  final annotations.  

\noindent \textbf{Stage 3: negative category verification.} By the design of federated annotation~\cite{lvis}, not all object categories are handled in each image. 
However, for evaluation purpose for each image we would need to collect a set of negative categories, defined as categories that do not appear in the image.
To operationalize this, we randomly sample several categories from the vocabulary $\mathcal{V}$ as the candidates of the negative categories, and ask annotators to verify. We remove a candidate category from the negative set if any annotator flags any corresponding object instance in the image. Finally, we get a set of negative categories per image.

\section{Benchmark Tasks on EgoObjects}
We introduce 4 benchmark tasks on EgoObjects, starting with a novel instance-level object detection task, which has been under-explored due to the lack of object ID annotations on individual objects captured in various conditions in existing datasets. We further present 2 novel continual learning object detection tasks, which are newly enabled by EgoObjects. Finally, we assess the performance of classical category-level object detection models on EgoObjects.

\subsection{Instance-Level Detection}
% \vspace{-0.2cm}
\label{sec:benchmark:inst_det}
In the applications of egocentric object understanding in AR and robotics, we are often confronted with the situation where the model is presented with few examples of object instances unseen during training and needs to detect those novel objects on-the-fly. Inspired by this, we introduce the instance-level detection below, and present two models, including a novel target aware- and a baseline target agnostic instance detector.

\begin{figure}[t!]
% \vspace{-0.1cm}
    \centering
    \includegraphics[width=1.01\columnwidth]{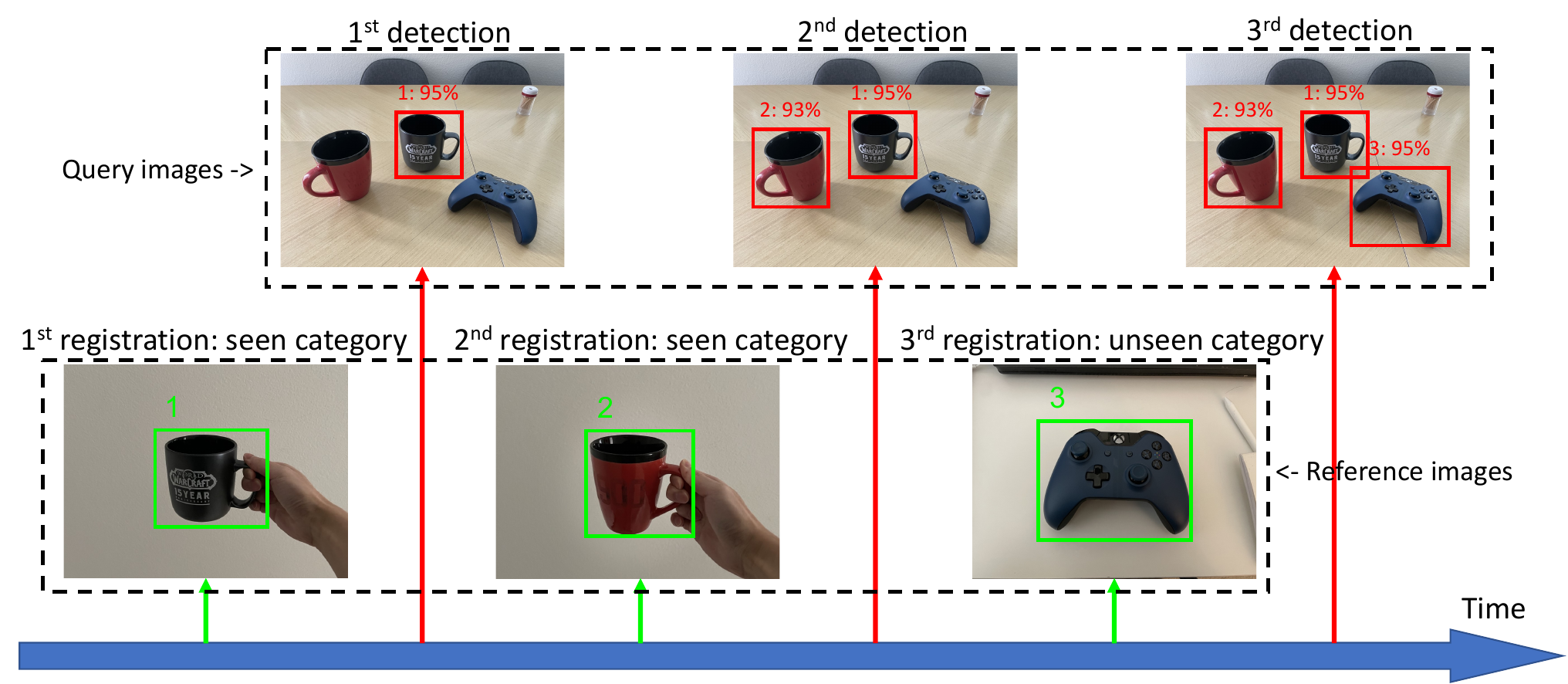}
    % \vspace{-0.3cm}
    \caption{\small \it \textbf{Instance detection at inference time}. 
    Continuously registering more targets leads to more detected objects, while previously registered targets are not forgotten. 
    The targets can be from either seen (target 1 and 2) or unseen (target 3) categories.}
    \label{fig:inst_det}
% \vspace{-0.5cm}
\end{figure}

% \vspace{-0.2cm}
\subsubsection{Task Specification}

At training time, the model has access to instance-level annotations of objects captured under diverse conditions. At inference time, the user can use the model to register a novel target instance $T$, regardless of whether its category is seen during training, by providing one or more 2D bounding boxes on reference images containing the target instance. After that, on a query image $I$, the detector needs to predict the bounding box of $T$ with $T$'s ID, or no box if $T$ is absent. 
To simulate the realistic setup where model back-propagation is difficult for deployed models,  we \emph{disallow} model fine-tuning on the target object instance annotations. The model is required to allow the user to continuously register more model targets, and all registered targets should be considered during detection. Figure \ref{fig:inst_det} contains an example where the user registers 3 targets sequentially and the model gradually detects more target objects in the same image.

\noindent \textbf{Dataset split.} We divide the dataset into 4 splits: {\tt train/target/val/test}. The {\tt train} split contains 9.6k instances with a total of 450k annotations from 79k images. The {\tt target}, {\tt val}, {\tt test} splits share the remaining 4.1k instances which do not appear in the {\tt train} images, and their categories can also be unseen during training.

In the {\tt target} split, there is a single reference image and one annotation for each instance. The {\tt val} and {\tt test} splits have 5.7K and 29.5K images with 3.8K and 4.1K instances, respectively.

\noindent \textbf{Evaluation protocols.} Under various IoU thresholds, we report Average Precision (AP) metrics, which are averaged across instances. 
Furthermore, we break down the metrics into two buckets for object instances from categories seen and unseen during training to assess the model capability of generalizing to instances from novel categories.

\subsubsection{Target-aware Instance Detector}

\begin{figure}[t!]
% \vspace{-0.5cm}
    \centering
    \includegraphics[width=1.01\columnwidth]{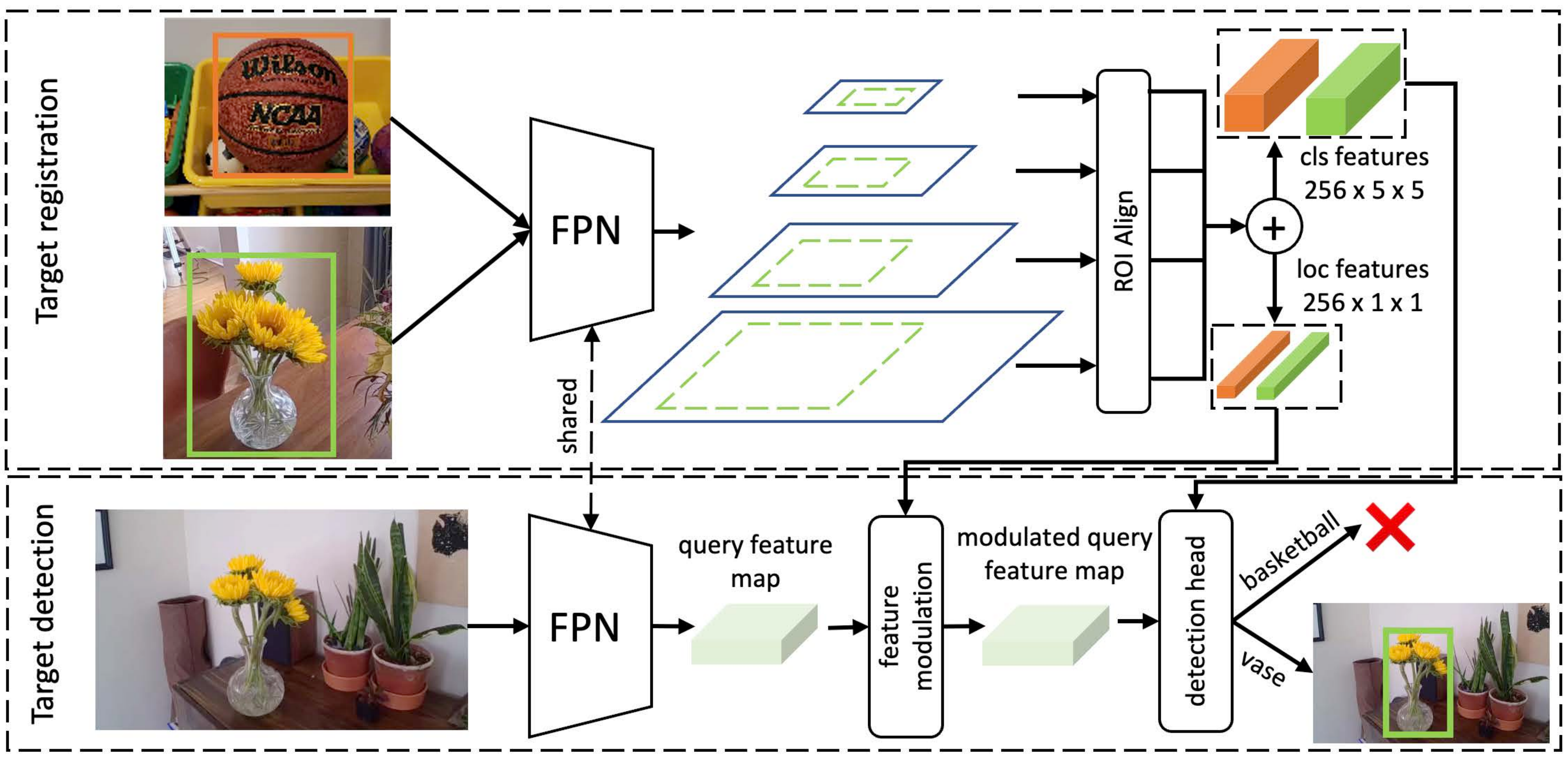}
    % \vspace{-0.3cm}
    \caption{\small \it \textbf{Architecture of target-aware instance detector \tadet}. \textbf{Top}: in target registration, localization and classification feature for each target are generated. \textbf{Bottom}: during target detection, the model predicts 1 bounding box per target and computes a confidence score to decide whether the prediction should be rejected via thresholding.}
    \label{fig:target_aware_inst_det}
% \vspace{-0.5cm}
\end{figure}

We propose an instance-level object detector aware of target objects during object localization, and refer to it as \emph{Target-Aware Instance Detector} (\tadet). It supports 2 modes, namely target registration and target detection (Figure~\ref{fig:target_aware_inst_det}).

\noindent \textbf{Target registration}. To register a new target object, we feed the reference image into a ResNet based FPN backbone~\cite{lin-fpn2017}, generate target features at different pyramid levels by using ROIAlign operator~\cite{mask-rcnn} according to the target bounding box annotation, and average them over pyramid levels. Target features of two different resolutions are obtained from ROIAlign. $T^{loc}$ feature of resolution $1\times1$ is used to efficiently localize the bounding box. $T^{cls}$ feature has higher resolution $S\times S$ ($S=$ 5 by default), and will be used to compute a confidence score of classifying the localized target object. If several reference images per target are provided, the target features are averaged.

\noindent \textbf{Target detection}. At detection time, \tadetspace use the same FPN backbone to extract query image feature map $F$ of size $C\times H \times W$ where $C$ denotes feature channels, and $\{H, W\}$ feature map size. A feature modulation block will transform $F$ according to target localization feature $T^{loc}$ of size $C\times 1 \times 1$, which attenuates the features in regions where the target object is less likely to appear. The detection head takes as input the modulated query feature map, and processes it using a \textit{Score} module, which consists of 4 convolutional layers with ReLU activations, to gradually reduce the channels from 256 to 1. The resulting score map is normalized by a \textit{Softmax} operation, and the target object center $(C_y, C_x)$ is predicted as the weighted sum of spatial coordinates according to the normalized score map.

% \begin{align*}
% \vspace{-0.3cm}
\begin{equation}
\small
\begin{aligned}
F^{mod} = (T^{loc} \circledast F) \odot F & \\
P = \textup{Softmax}( \textup{Score}(F^{mod}).\textup{reshape}(-1)) & \\
Y^{g} = \textup{ls}(0, H-1, \textup{steps}=H).\textup{view}(H, 1).\textup{repeat}(1, W) & \\
X^{g} = \textup{ls}(0, W-1, \textup{steps}=W).\textup{view}(1, W).\textup{repeat}(H, 1) & \\
C_{y} = \textup{sum}( P \odot Y^{g}.\textup{reshape}(-1)) & \\
C_{x} = \textup{sum}( P \odot X^{g}.\textup{reshape}(-1)) &
\end{aligned}
\label{eqn:target detection}
\end{equation}
% \end{align*}
% \vspace{-0.2cm}

\noindent where $\circledast$ denotes convolution, $\odot$ element-wise multiplication and $\textup{ls } \textup{torch.linspace}$. To refine object center and predict target object size, we sample a target feature at $(C_y, C_x)$ in $F$ via bilinear interpolation, and employ a 3-layer MLP with hidden dimension $256$ to predict the spatial offset $(\delta C_y, \delta C_x)$ and target object size $(S_y, S_x)$ with a ReLU activation. After predicting target object box, we use ROIAlign to sample a spatial feature of the resolution $S \times S$ in $F$, and compute its dot product with $T^{cls}$ using a sigmoid activation function as the box confidence score.

\noindent \textbf{Model training}.  During training, we sample three images for each instance: one reference image containing the instance, one positive image containing the instance captured in a different setting, and one negative image that does not contain the instance. In positive image, we consider both bounding box localization loss and classification loss. For localization loss, we use a linear combination of $L_1$ loss and generalized IoU loss~\cite{giou}. For classification loss, we use the binary cross entropy loss between the predicted box confidence score and groundtruth box label, which is positive when IoU is above $IoU^{pos}$, negative when IoU is below $IoU^{neg}$ and ignored otherwise. By default, ($IoU^{pos}$, $IoU^{neg}$) = $(0.7, 0.3)$. In negative image, only the classification loss is used and groundtruth label is negative. See more studies in the supplement.

\subsubsection{Baseline Target-agnostic Instance Detector}
We also consider a simple baseline approach \textit{RPN+SFNet} which consists of a Region Proposal Network (RPN) \cite{faster-rcnn} for object localization and a SFNet model~\cite{sphereface2}, commonly used in metric learning, for object classification. We briefly review its target registration, detection and model training below, and include more details in the supplement.

\noindent \textbf{Target registration}. We crop the target object from reference images and feed it through the SFNet model to obtain the target object feature, which is then added to an index of target object features. %to complete registration for the object.

\noindent \textbf{Target detection}. For a given query image, the RPN generates a large number of object proposals \textit{agnostic} to the target objects in the index. Each object proposal is cropped from the query image and fed into the SFNet model to extract the feature. These object features are then matched against all the added target features in the index. We pick the target object in the index with the highest matching score. The final confidence score of an object proposal against the top target object is the product of RPN object proposal confidence score and its matching score with the target object.

\noindent \textbf{Model training}. The RPN is trained on the train split using all bounding box annotations. The SFNet model is trained with SphereFace2 \cite{sphereface2} loss function using all instance-level annotations, which encourages small distances between features of multiple views of the same instance, and large distances for features of different instances.

\begin{table}
% \vspace{-0.5cm}
\centering
\setlength{\tabcolsep}{2pt}
% \small
\resizebox{1.01\columnwidth}{!}{
\begin{tabular}{c|c|cccc|cccc}
 &  & \multicolumn{4}{c|}{val} & \multicolumn{4}{c}{test} \\
backbone & method & AP & AP50 & AP50$_{se}$ & AP50$_{un}$ & AP & AP50 & AP50$_{se}$ & AP50$_{un}$ \\ \shline
\multirow{2}{*}{R50} & RPN+SFNet & 17.8 & 29.0 & 29.1 & 19.8 & 15.7 & 25.4 & 25.5 & 16.8 \\
 & \tadet & 18.7 & 35.0 & 35.0 & 21.7 & 18.5 & 35.2 & 35.2 & 24.8 \\ \hline
\multirow{2}{*}{R101} & RPN+SFNet & 19.3 & 32.0 & 32.0 & 22.3 & 17.0 & 27.7 & 27.8 & 20.0 \\
 & \tadet & 22.6 & 37.9 & 38.0 & 28.5 & 21.9 & 37.9 & 38.0 & 26.4 \\ \hline
\end{tabular}
}
% \vspace{-0.2cm}
\caption{\small \it \textbf{Instance-level detection benchmarking results on EgoObjects.} The proposed \textup{\tadet} model significantly outperforms the baseline \textup{RPN + SFNet} approach. AP50$_{se}$ and AP50$_{un}$ are computed for instances with categories seen and unseen during training. On the more challenging test split with more targets object instances, \tadetspace can maintain the performance whereas \textup{RPN + SFNet} baseline has a significant performance drop.  R50 and R101 denote ResNet-50/101 backbones.
% \fyx{Add some comments on our hypothesis}
}
\label{tab:inst_det_res}
% \vspace{-0.5cm}

\end{table}

\subsubsection{Results}

The benchmark results of both models are presented in Table \ref{tab:inst_det_res}. 
For both approaches \textit{\tadet} and \textit{RPN+SFNet}, we build models with ResNet-50 and ResNet-101 backbones. There are several intriguing observations. First, \textit{\tadet} substantially outperforms the \textit{RPN+SFNet} on all metrics. For example, the gains in AP50 are large ($+6\%$ on val and $+10\%$ on test split). We attribute this to the design that \textit{\tadet} localizes the target object by using query image feature maps modulated by the target feature, and does not rely on the target-agnostic RPN to generate object proposals. 

Second, the best \textit{\tadet} model with ResNet-101 backbone only achieves less than $23\%$ AP on both val and test split which have around 4K novel instances each, indicating the unique challenges in the large-scale instance-level object detection, such as large changes in viewing direction, lighting, background and distance as well as less distinguishable appearance between instances from the same category. See more examples in the supplement.

Third, there are significant gaps between AP$_{se}$ and AP$_{un}$, reflecting the challenges in generalizing the models to detect instances from categories unseen during training.

\subsection{Continual Learning}

Existing continual learning (CL) approaches often tackle object classification problem while continual learning object detection task is not well explored due to the lack of a large-scale dataset that is annotated with instance- and category-level labels, and contains individual objects in multiple images captured under diverse conditions. We introduce 2 novel CL tasks, namely \emph{CL instance detection} and \emph{CL category detection}, on a subset of EgoObjects which contains 100K images with 250K box annotations for 1.1K main object instances from 277 categories. There are 3.4K and 3.1K instances in the train- and test set, respectively.

\noindent \textbf{CL Instance Detection.}
In this task, we simulate the setting when a system continuously encounters new batches of instance detection training data, where each batch indexed at $i$ is called an experience $E_i$. In $E_i$, each image only carries the annotation of its main object with instance ID being the class. The system can only access data in the latest experience, which means no access to previous experiences apart from the use of a limited replay memory. Previous experiences share no common main object instances with later experiences, which makes it a \emph{Class-Incremental Learning} setup. We evenly split $1110$ main instances into 5 experiences, hence 222 instances per experience. 
For evaluation, the system is benchmarked after each experience on a fixed testing set with all the main instances, making it a $1110$-way detection problem. The evaluation metric for each experience is mean average precision (mAP). The final performance is the averaged metric across all the experiences.

\noindent \textbf{CL Category Detection.}
In this task, the goal is to predict the object category labels instead of instance IDs. We create 5 experiences by applying the class-incremental ordering on the 277 categories of the main object instances. 
Additionally, we also include the annotations of secondary objects, which makes it a \emph{Data-Incremental Learning} setup, i.e. previous experiences share no common images or annotations with later ones. This differentiates our task from other CL object detection tasks focusing on annotation incrementality, where the same images are repeatedly encountered in successive experiences but with a different set of annotations (usually class-incremental). We believe our task provides a more realistic setting.  The evaluation metric for each experience is also mAP.

\noindent \textbf{Results.} We benchmark the methods from the top submissions of the 3rd CLVision workshop challenge \cite{clworkshop} and report the results on above CL tasks in Table \ref{tab:cl_det}. In general, these submissions build upon classic 1-stage/2-stage detectors and adopt techniques to mitigate catastrophic forgetting of early experiences when trained on later experiences, such as sampling data from previous experiences cached in a replay buffer, and distilling models from early experiences to the model for the current experience. 
However, these winning methods still have limitations. They treat the instance detection as a close-set problem same as the category detection, which cannot scale up to flexibly accommodate more instances. Additionally, there is no change to the detector architecture to better tailor to the CL tasks. See more discussions in the supplement.

\begin{table}[t!]
% \vspace{-0.7cm}
\centering
\setlength\tabcolsep{3pt}
% \small
\resizebox{\columnwidth}{!}{
\begin{tabular}{l|cccccc|cccccc} 
% \hline
& \multicolumn{6}{c|}{CL Instance Detection} & \multicolumn{6}{c}{CL Category Detection} \\
rank & $E_0$ & $E_1$ & $E_2$ & $E_3$ & $E_4$ & $EAP$ & $E_0$ & $E_1$ & $E_2$ & $E_3$ & $E_4$ & $EAP$ \\ 
\shline
1st & 23.3 & 39.5 & 54.6 & 70.2 & 85.6 & 54.7 & 30.6 & 47.2 & 58.1 & 67.5 & 76.2 & 55.9 \\
2nd & 15.1 & 30.4 & 45.5 & 60.8 & 75.4 & 45.4 & 28.4 & 44.7 & 57.6 & 67.9 & 78.2 & 55.4 \\
3rd & 14.7 & 29.1 & 42.3 & 55.4 & 66.9 & 41.7 & 19.5 & 34.5 & 43.9 & 52.7 & 61.5 & 42.4 \\
\hline
\end{tabular}
}
% \vspace{-0.3cm}
\caption{\small \it \textbf{CL detection benchmarks on EgoObjects.} We report detection accuracy (mAP) after each experience and final Experience Average Precision (EAP), which is the averaged mAP over experiences.}
\label{tab:cl_det}
% \vspace{-0.6cm}
\end{table}

\begin{table*}[t]
  \centering
  \small
  \resizebox{2.0\columnwidth}{!}{
  \begin{tabular}{l | c | c c c c c c c c c c | c c}
    \multicolumn{1}{c|}{} & \multicolumn{1}{c|}{} & \multicolumn{10}{c|}{val} & \multicolumn{2}{c}{test}
    \\
    \multicolumn{1}{c|}{method} & pretrain & AP & AP75 & AP50 & AP50$_l$ & AP50$_m$ & AP50$_s$ & AP50$_{bright}$ & AP50$_{dim}$ & AP50$_{simple}$ & AP50$_{busy}$ & AP & AP50
    \\
    \shline  % model path example: manifold://fai4ar_osu/tree/users/fanyix/fbl_runs/2022-12-21_11-10-57/e2e_train/model_final.pth
    FasterRCNN-R50 & IN1k & 20.5 &  21.8 &  32.9 &  41.4 &  31.2 &  18.1 & 35.6 & 32.3 & 42.3 & 32.7 & 20.1 & 32.6 % f397364884, f411007190, 2022-12-21_11-10-57
    \\
    FasterRCNN-R101 & IN1k & 22.5 & 23.6 & 37.9 & 47.0 & 35.8 & 20.8 & 40.7 & 37.5 & 49.2 & 37.1 & 21.6 & 36.6 % f397411470, f411006846, 2022-12-21_14-29-56
    \\
    \hline
    FCOS-R50 & IN1k & 23.1 & 25.0 & 31.4 & 40.2 & 28.1 & 16.9 & 33.2 & 31.7 & 40.0 & 31.4 & 22.4 & 30.6 % f404301338, f411006651, 2023-01-19_22-21-11
    \\
    FCOS-R101 & IN1k & 24.2 &  26.4 &  32.1 &  40.5 &  28.9 &  17.8 & 34.2 & 31.9 & 41.3 & 31.9 & 23.2 & 31.0 % f404302417, f411006517, 2023-01-19_22-36-49
    \\
    \hline
    DeformDETR-SwinT & IN22k & 29.1 &  31.6 &  37.7 &  47.0 &  34.3 &  22.5 & 39.8 & 37.9 & 48.3 & 37.3 & 28.6 & 37.1 % f403705495, f411006291, 2023-01-17_18-31-58
    \\
    DeformDETR-SwinS & IN22k & 32.0 &  34.4 &  41.0 &  52.1 &  37.5 &  21.9 & 42.8 & 41.7 & 52.8 & 39.8 & 31.1 & 40.3 % f414612989, f416098592, 2023-02-27_21-16-12
    \\ 
    DeformDETR-SwinB & IN22k & 32.2 & 34.8  &  41.7 &  51.0 &  39.3 &  23.8 & 43.8 & 41.7 & 53.0 & 40.2 & 31.1 & 40.4 % f404257724, f411005916, 2023-01-19_17-29-25
    \\
    DeformDETR-SwinL & IN22k & 33.5 &  36.3 &  42.6 &  53.7 &  39.5 &  25.6 & 44.7 & 42.9 & 53.2 & 42.2 & 32.6 & 41.8 % f403726038, f411005557, 2023-01-17_20-14-27
    \\
    \hline
  \end{tabular}}
  \caption{\small \it \textbf{Category detection benchmark on EgoObjects.} We benchmark three mainstream detectors in this table: FasterRCNN~\cite{faster-rcnn}, FCOS~\cite{tian-fcos2019}, and Deformable-DETR~\cite{zhu-deformabledetr2020}. All detectors are trained on {\tt train} split and tested on {\tt val} split. \{AP50$_l$, AP50$_m$, AP50$_s$\} are computed for different object sizes, \{AP50$_{bright}$, AP50$_{dim}$\} for different lighting conditions, and \{AP50$_{simple}$, AP50$_{busy}$\} for different background scenes. 
  }
  \label{table:catdet}
\end{table*}

\subsection{Category-Level Detection} EgoObjects also supports the classical category-level object detection task given the nearly 400 object categories in the current Pilot version.

\noindent\textbf{Evaluation protocols.} We use the same dataset splits as the instance-level detection task.   
In total, there are 447K/31K/164K object annotations from 368 categories in the {\tt train/val/test} split, respectively.  Due to its federated annotation process, we only penalize false positive predictions on an image if the predicted class is in the list of negative categories for that image.

\noindent\textbf{Benchmarking models}. We consider 3 types of performant object detectors. The first one is FasterRCNN~\cite{faster-rcnn}, which is a two-stage detector.  Next, we include the representative single-stage detector FCOS~\cite{tian-fcos2019} which skips the explicit proposal generation to accelerate the model inference. Finally, we also consider the recent transformer-based detectors (\ie, DETR~\cite{carion-detr2020}). Specifically, we adopt the Deformable-DETR~\cite{zhu-deformabledetr2020} due to its stable and fast training. For both FasterRCNN and FCOS, we use the ResNet50/101 backbone pretrained on ImageNet-1K, whereas for Deformable-DETR, we use the Swin-Transformers~\cite{liu-swin2021} backbone pretrained on ImageNet-22K.

\noindent \textbf{Results}. The results are presented in Table~\ref{table:catdet}. Notably, single stage FCOS models outperform two-stage FasterRCNN detectors particularly for high IOU threshold (\eg AP75), while DeformDETR-Swin models significantly outperform both types of CNN detectors at the cost of large model size and significantly more compute. However, even for the largest DeformDETR-SwinL model, its AP metrics on EgoObjects are still $10\%$ lower than its AP on LVIS $43.7\%$ reported in Table 3 of prior work~\cite{hdetr}. We hypothesize due to the egocentric view and its data capture setting, EgoObjects contains larger variations in background, viewpoints, lighting and object distances, which together render it more difficult even for category-level detection.
We also implemented the metrics of different buckets on experimental conditions (e.g. object scale, lighting, background complexity) in our evaluation API. We observe model's performance is lower under the more challenging conditions (small scale, dim lighting, busy background).

% \vspace{-0.3cm}
\section{Conclusions}
We present EgoObjects, a large-scale egocentric dataset containing tens of thousands of videos, and more than half million object annotations. By design, it captures the same object under diverse conditions while annotating it with both category label and consistent object IDs in multiple images.
To stimulate the egocentric object understanding research on it, we introduce 4 tasks and also provide the benchmarking results of various models, including a novel target-aware instance-level detector which largely outperforms an off-the-shelf baseline based on RPN and SFNet.

{\small
\bibliographystyle{ieee_fullname}
\bibliography{egbib}
}

\clearpage

\appendix
\appendixpage
\addappheadtotoc

% \begin{appendices}

\section{Additional details for data collection}

Given 4 variables including object distance, camera motion, background complexity, and lighting, participants are instructed to collect 10 videos of each main object according to 10 predefined configurations, which is presented in Table \ref{tab:collection_setting}. Figure \ref{fig:data_intro} shows representative frames of these 10 videos featuring one main object - pressure cooker.

% Moreover, we attach the full hierarchical structure of the object category taxonomy in {\tt Object Taxonomy.pdf}, a 7-layer tree structure with a total of 638 leaf nodes.

\section{Additional details for data annotation}

\begin{figure*}
    \centering
    \begin{subfigure}[t]{\columnwidth}
        \includegraphics[width=\textwidth]{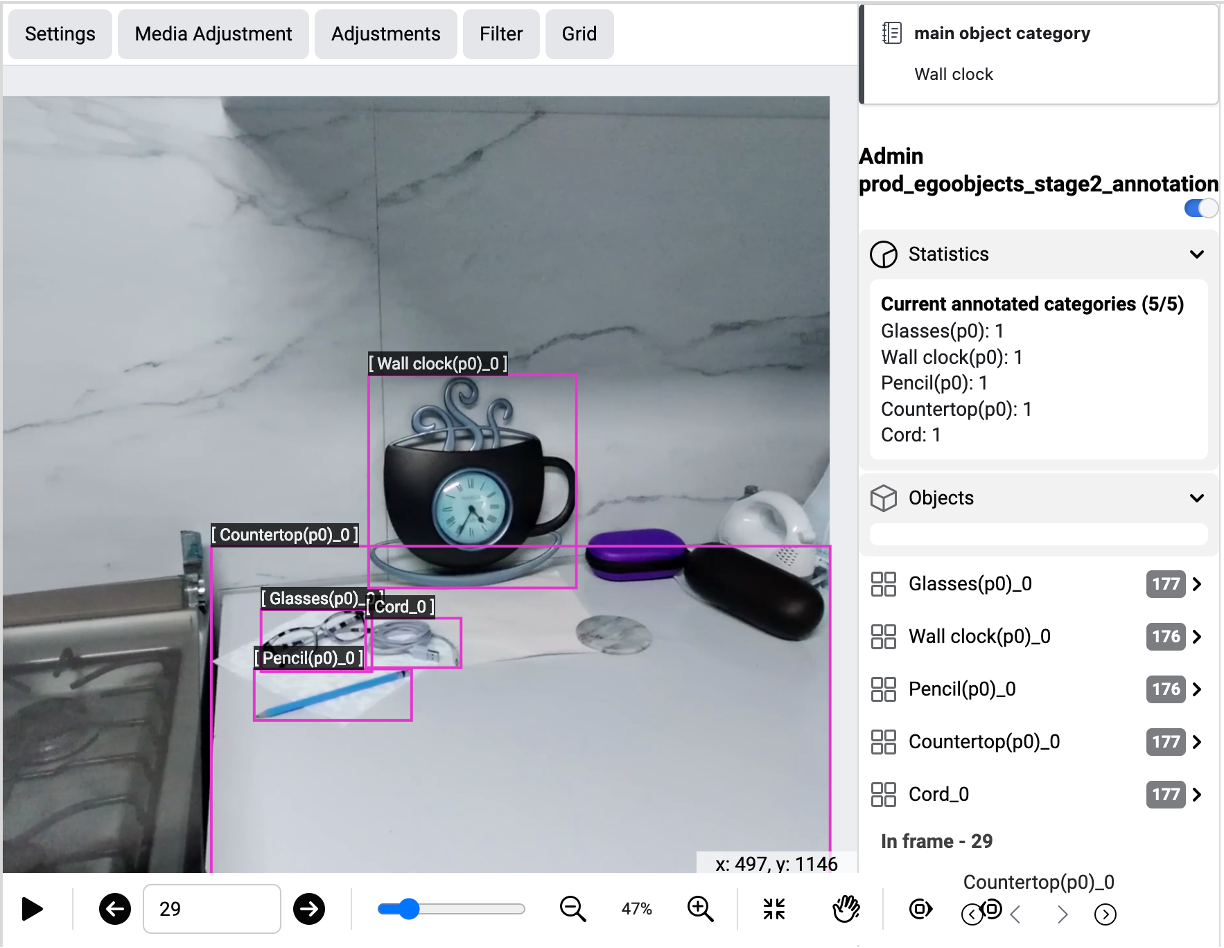}
        \caption{\small \it Annotation UI for stage 1\&2.}
        \label{fig:anno_stage2}
    \end{subfigure}
    \quad
    \begin{subfigure}[t]{\columnwidth}
        \includegraphics[width=\textwidth]{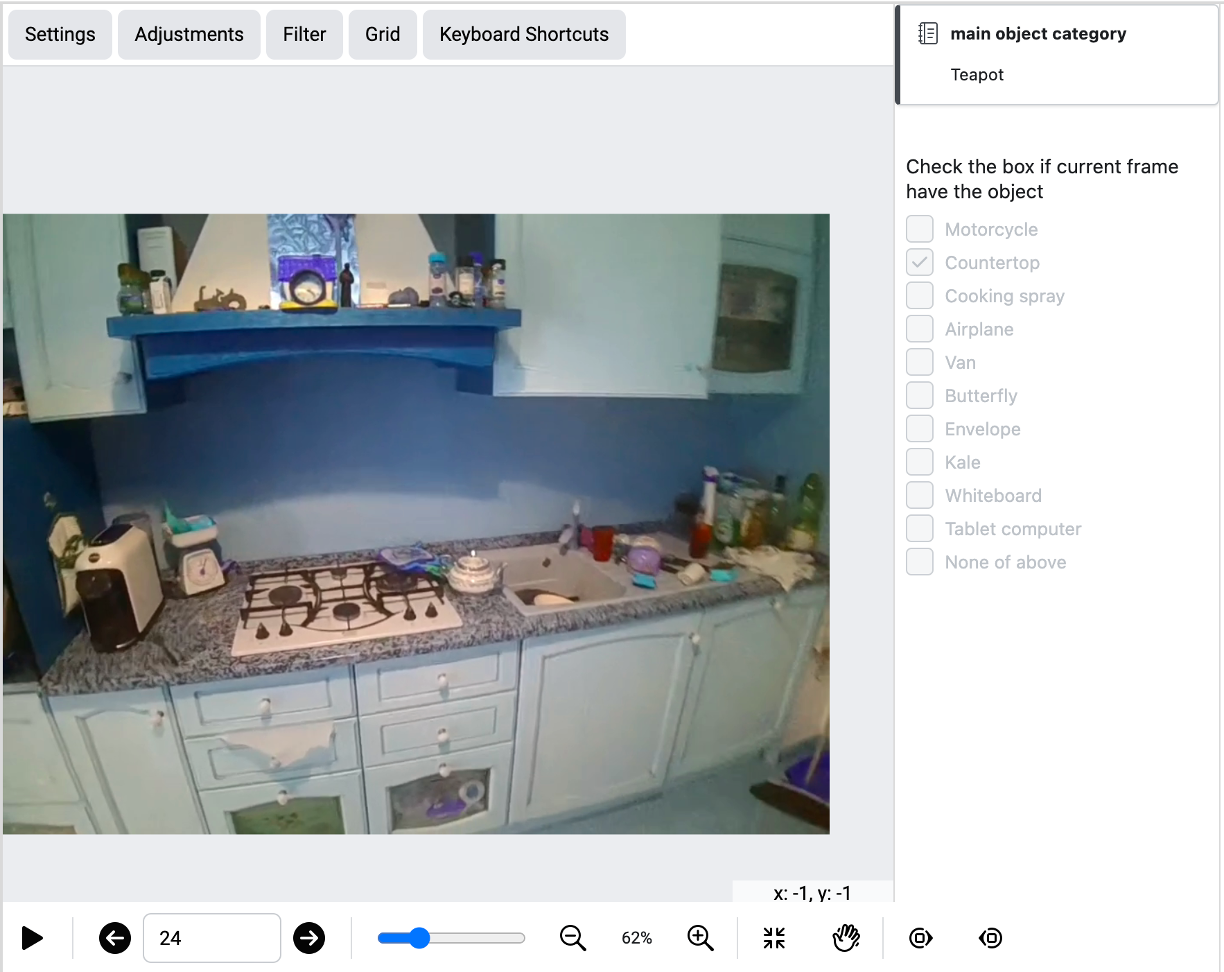}
        \caption{\small \it Annotation UI for stage 3.}
        \label{fig:anno_stage3}
    \end{subfigure}
    \caption{\small \it \textbf{Annotation tools for multi-stage process}. Stage 1\&2 is done in a per-frame object bounding box interface (left). Stage 3 is done in a per-frame attribute verification interface (right).}
    \label{fig:multi-stage-annotation}
\end{figure*}

We demonstrate the annotation tools for our multi-stage annotation process in Figure \ref{fig:multi-stage-annotation}. Both tools have video player as the back end. We enqueue all 10 videos of the same main object into a single job to ensure label consistency.

For stage 1 (category discovery) and stage 2 (exhaustive instance labeling), we use a per-frame object bounding box interface (Figure \ref{fig:anno_stage2} with some panels on the right for additional information. Annotators are asked to label a track for each identified instance across the frames by drawing bounding boxes and assigning them consistent category label and instance ID. Annotators can consult the ``main object category'' panel and the ``statistics'' panel for the number of annotated categories and number of instances for each category in the current frame. Annotators can also quickly jump to other frames of the track of an instance by navigating through the ``Objects'' panel.

For stage 3 (negative category verification), we use a per-frame attribute classification interface (Figure \ref{fig:anno_stage3}). On the right panel there is a list of categories with check boxes. Annotators are asked to check for each category whether there is any instance of that category in the image. Categories not ticked are verified negative categories for the image.

Figure \ref{fig:data_intro} shows representative annotations of both the main object and the surrounding secondary objects.

\section{Ablation studies on TA-IDet}

\begin{table}
\centering
% \small
\setlength\tabcolsep{5pt}
\resizebox{\columnwidth}{!}{
\begin{tabular}{c|cccc}
% \hline \hline
video ID & distance & camera motion & background & lighting \\
\shline
01       & near            & horizontal    & simple     & bright   \\
02       & medium          & horizontal    & simple     & bright   \\
03       & near            & horizontal    & simple     & dim      \\
04       & medium          & horizontal    & busy       & bright   \\
05       & far             & horizontal    & busy       & bright   \\
06       & medium          & vertical      & busy       & bright   \\
07       & medium          & combined      & busy       & bright   \\
08       & near            & horizontal    & busy       & dim      \\
09       & medium          & horizontal    & busy       & dim      \\
10       & far             & horizontal    & busy       & dim      \\
\hline
\end{tabular}
}
\caption{\small \it Different settings of video condition variables for 10 videos of each main object.}
\label{tab:collection_setting}
\end{table}

\begin{table}
\centering
\setlength\tabcolsep{5pt}
\resizebox{\columnwidth}{!}{
\begin{tabular}{ccc|lll}
\multicolumn{3}{c|}{loss weight} & \multicolumn{3}{c}{val} \\
index & positive & negative & AP & AP50 & AP75 \\ \shline
\xmark & \cmark & \xmark & 12.9 & 25.3 & 11.7 \\
\xmark & \cmark & \cmark & 13.6$_{+0.7}$ & 27.2$_{+1.9}$ & 12.1$_{+0.4}$  \\
\cmark & \cmark & \xmark & 17.4$_{+4.5}$ & 31.4$_{+6.1}$ & 17.1$_{+5.4}$ \\
\cmark & \cmark & \cmark & 18.7$_{+5.8}$ & 35.0$_{+9.7}$ & 17.7$_{+6.0}$ \\ \hline
\end{tabular}
}
\caption{\small \it \textbf{Ablation study on the training losses of TA-IDet.} We train TA-IDet with different combinations of losses and report the performance on the EgoObjects {\tt val} split. Backbone is ResNet-50. }
\label{tab:loss_weight}
\end{table}

At model training, we can compute three losses: positive loss, index loss, and negative loss.
The positive loss is the instance detection loss in the positive image, which contains the same object instance in a different view as in the index image.
The index loss is the instance detection loss in the index image. The supervision signal is the model should be able to localize and recognize the same object instance in the original index image.
The negative loss refers to the classification loss in the negative image. The negative image does not contain the same object instance in the index image. Regardless of what object bounding box is predicted by the instance localization module, the object confidence score is minimized.
We study the effects of the index loss and the negative loss by setting their loss weights to zero and compare with model trained with all the losses. In Table \ref{tab:loss_weight}, we confirm the highest accuracy is achieved when all three losses are used.

Additionally, we study the effects of different $T^{cls}$ feature resolutions $S\times S$ in Table \ref{tab:feat_cls_size}. We find $S=5$ produces the best accuracy.

\begin{table}
\centering
% \setlength\tabcolsep{5pt}
% \resizebox{\columnwidth}{!}{
\begin{tabular}{c|ccc}
 & \multicolumn{3}{c}{val} \\
 $S$ & AP & AP50 & AP75 \\ \shline
1 & 11.3 & 20.4 & 11.3 \\
3 & 18.5 & 34.6 & 17.6  \\
\textbf{5} & \textbf{18.7} & \textbf{35.0} & \textbf{17.7} \\
7 & 18.3 & 34.3 & 17.4 \\ \hline
\end{tabular}
% }
\caption{\small \it \textbf{Ablation study on the $T^{cls}$ feature resolution of TA-IDet.} We train TA-IDet with different $T^{cls}$ feature resolution $S$ and report the performance on the EgoObjects {\tt val} split. Backbone is ResNet-50. }
\label{tab:feat_cls_size}
\end{table}

\section{Implementation details of target-agnostic instance detector}

\begin{figure}
    \centering
    \includegraphics[width=\columnwidth]{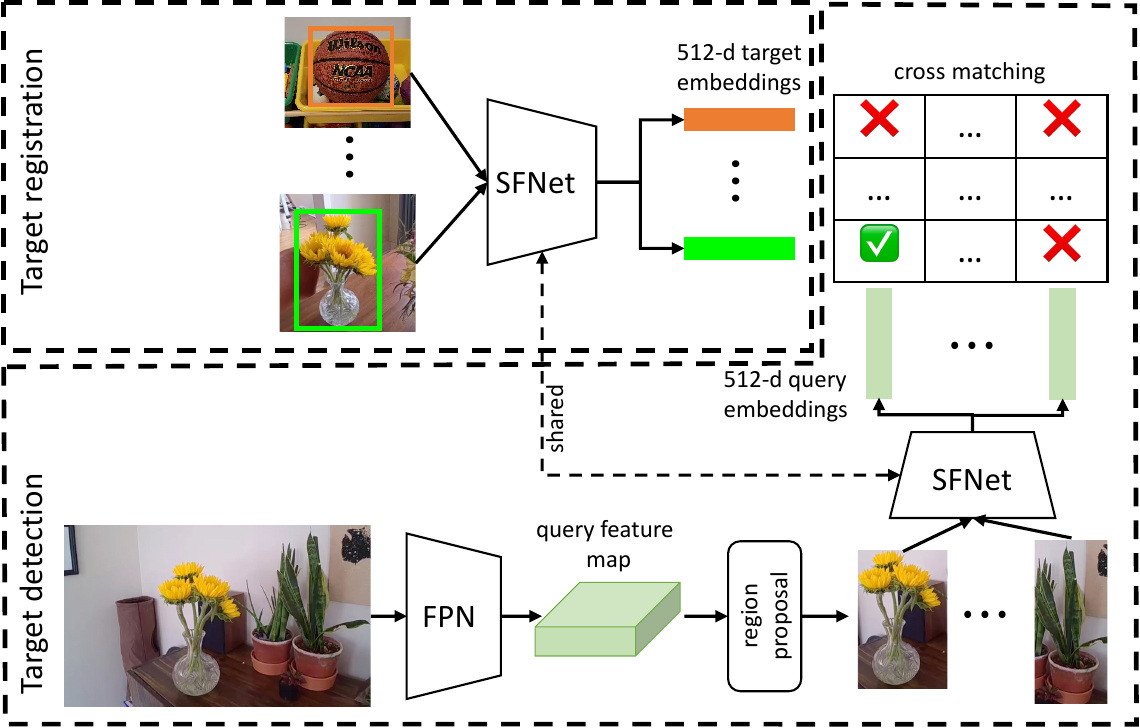}
    \caption{\small \it \textbf{Architecture of target-agnostic instance detector.} \textbf{Top:} in target registration, target embeddings are generated from SFNet \cite{sphereface2}. \textbf{Bottom:} during target detection, the model generates region proposals and their corresponding query embeddings, which are cross-matched with target embeddings via cosine similarity. }
    \label{fig:target_agnostic_det}
\end{figure}

We propose a simple baseline approach \emph{RPN+SFNet} as the target-agnostic instance detector. It consists of a Region Proposal Network (RPN) \cite{faster-rcnn} for object localization and a SFNet model \cite{sphereface2}, commonly used in metric learning, for object classification. It supports two modes, namely target registration and target detection, see Figure \ref{fig:target_agnostic_det}.

\noindent \textbf{Target registration.} To register a new target instance, we crop the instance out from the reference image and resize the crop to a fixed size ($112 \times 112$). Then a 512-dimensional embedding is extracted by passing the crop through the SFNet and added to an index of target object embeddings. If several reference images per target are provided, the target embedding is averaged.

\noindent \textbf{Target detection.} At detection time, we feed the query image into RPN to generate several instance proposals agnostic to the targets in the index. Each object proposal is cropped from the query image and also resized to $112 \times 112$. Then we use the same SFNet as in target registration to extract 512-dimensional query embedding for each proposal. All the query embeddings are matched against all the target embeddings in the index by computing the cosine similarity. We pick the target instance in the index with the highest cosine similarity as the matched one for each proposal. We also apply a threshold to the matching score for rejecting low confident matches.

\noindent \textbf{Model training.} During training, the RPN is trained on the {\tt train} split with all the object bounding boxes in a instance-agnostic way. For the SFNet, we build a classification dataset by pre-cropping all the instances from {\tt train} images and assign them labels as the instance IDs, so that each instance has several samples of multiple views. The SFNet model is trained on this classification dataset with the SphereFace2 \cite{sphereface2} loss, which encourages small distances between embeddings of multiple views for the same instance and large distances for embeddings of different instances. SphereFace2 is the state-of-the-art deep face recognition method. It constructs a novel binary classification framework to align the training objective with open-set verification, outperforming other metric learning losses on recognizing unseen objects. We follow the best practices of \cite{sphereface2} and adopt its hyperparameters in our model training.

% \begin{equation}
%     sf2 loss = ...
% \end{equation}

\section{Dataset challenges}

\begin{figure*}
    \centering
    \includegraphics[width=2\columnwidth]{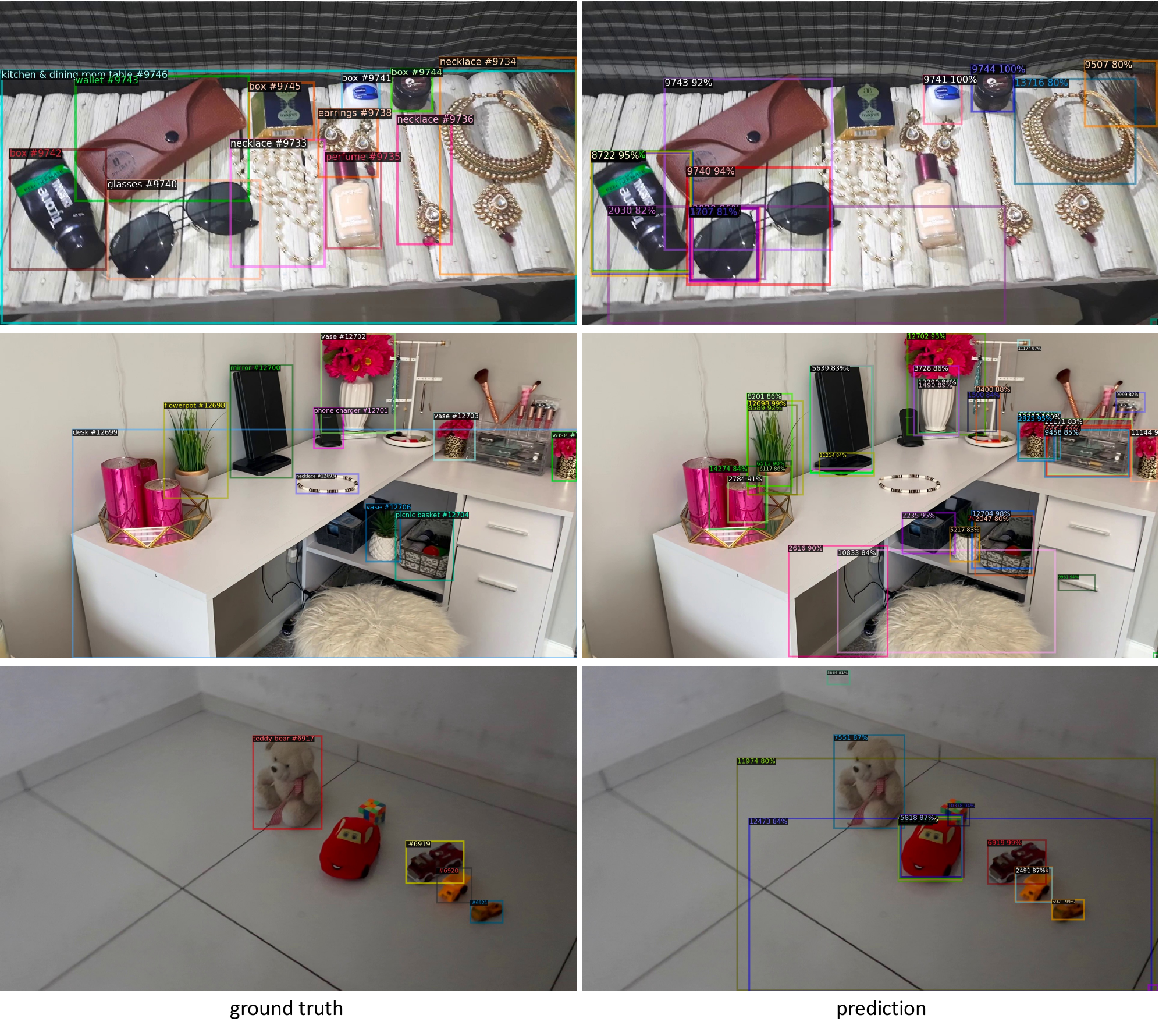}
    \caption{\small \it Visualization of ground truths (left) and predictions (right) for \textbf{instance-level detection}. Predictions are produced from a TA-IDet-R101 model on the {\tt val} split. The labels on top of boxes contain instance IDs. Best viewed digitally.}
    \label{fig:vis_ins_det}
\end{figure*}

\begin{figure*}
    \centering
    \includegraphics[width=2\columnwidth]{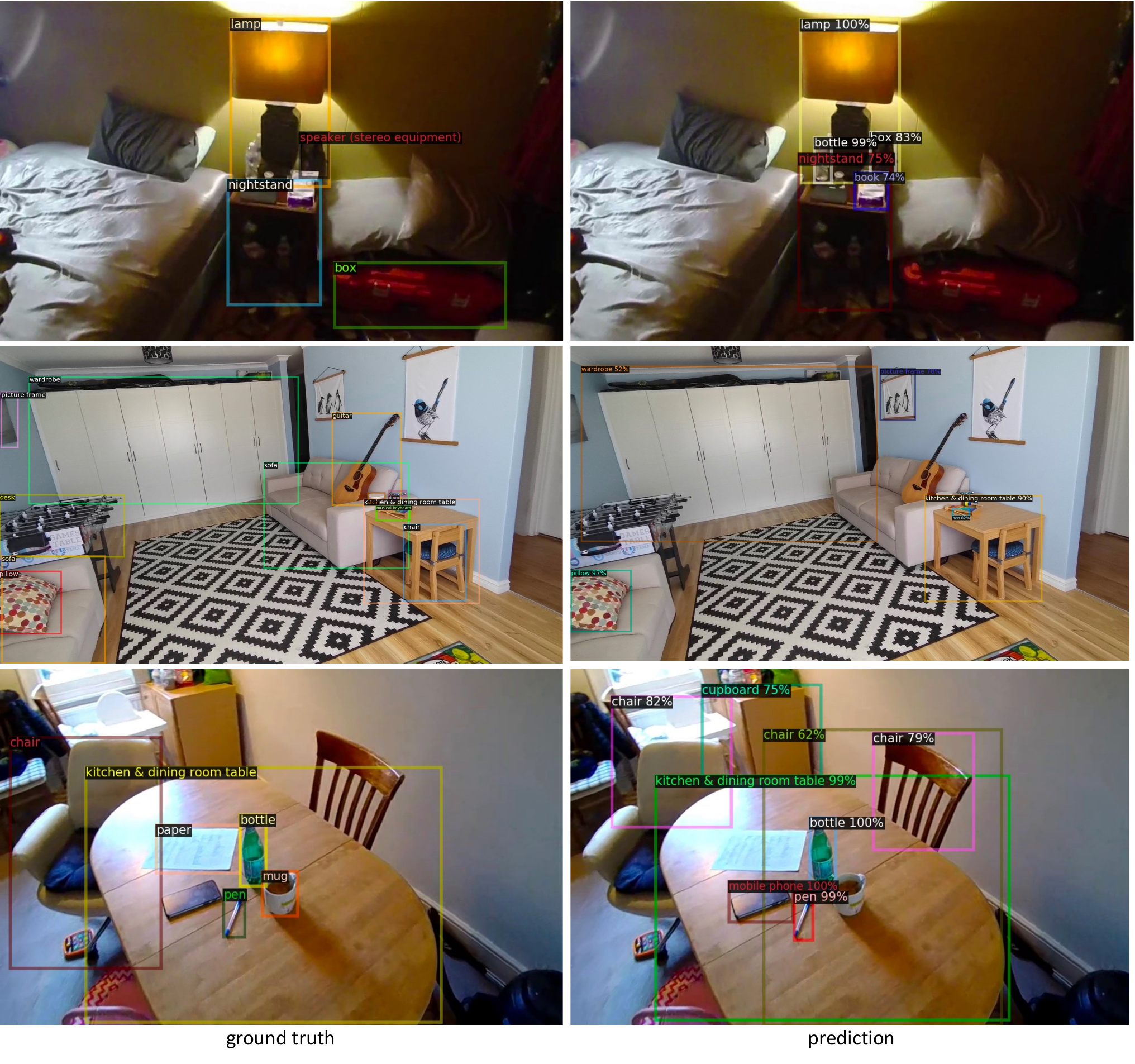}
    \caption{\small \it Visualization of ground truths (left) and predictions (right) for \textbf{category-level detection}. Predictions are produced from a FasterRCNN-R101 model on the {\tt val} split. Best viewed digitally.}
    \label{fig:vis_cat_det}
\end{figure*}

We present several qualitative results of instance-level detection in Figure \ref{fig:vis_ins_det} and category-level detection in Figure \ref{fig:vis_cat_det} to demonstrate the challenges with tasks defined on EgoObjects. For instance-level detection, one dominant challenge is the massive number of instances. An ideal instance detector should be able to output high confidence score for each target while suppressing the confidence scores of other targets. But we observed the prediction of multiple instances with high confidence on a single object, even from the best TA-IDet-R101 model. For category-level detection, challenges mainly come from the diverse capture conditions like far distance, cluttered background, dim lighting, extreme view angles, etc. 

\section{New splits for instance detection with seen/unseen categories}
In the benchmark tasks of the main text (Section 4), we designed a set of data splits for benchmarking both category-level detection and instance-level detection uniformly, which maximizes the category overlap for training and testing, resulting in only 25 instances in the evaluation with unseen categories. We refer to this set of splits as \textbf{Unified} splits. To further study the challenges in generalizing instance detection models to instances with unseen categories, we created a new set of data splits, named \textbf{InstDet} splits, to have more instances with unseen categories.

For the InstDet splits, we also divide the dataset into 4 subsets: {\tt train/target/val/test}. The {\tt train} split contains 12.2k instances with a total of 543k annotations from 104k images. The {\tt target}, {\tt val}, {\tt test} splits share the remaining 1.5k instances which do not appear in the {\tt train} images, and their categories can also be unseen during training.
Among those 131 instances have \emph{unseen} categories. 
In the {\tt target} split, there is a single reference image and one annotation for each instance. The {\tt val} and {\tt test} splits have 2.5K and 7.8K images respectively. We compare the new InstDet splits with Unified splits in Table \ref{tab:splits_comparison}.

Table \ref{tab:inst_det} reports the results. There are clear gaps between AP$_{se}$ and AP$_{un}$ on both {\tt val} and {\tt test} splits, reflecting the challenges in generalizing the models to unseen categories. Models with R101 backbone tend to have larger gaps compared to models with R50 backbone, indicating the trend of overfitting to seen categories. Especially for RPN+SFNet baseline, the larger backbone does not bring as much gain on AP50$_{un}$ as the TA-IDet approach. On the other hand, TA-IDet robustly outperforms RPN+SFNet on AP50$_{un}$ in all the settings, indicating its superiority of generalizing to instances with unseen categories.

\begin{table}
    \centering
    \setlength\tabcolsep{3pt}
    \resizebox{\columnwidth}{!}{
    \begin{tabular}{c|cccccc}
       splits & \#train img & \#val img & \#test img & \#train inst & \#eval inst$_{sc}$ & \#eval inst$_{uc}$ \\ \shline
        Unified & 79K & 5.7K & 29.5K & 9.6K & 4.1K & 25 \\
        InstDet & 104K & 2.5K & 7.8K & 12.2K & 1346 & 131 \\ \hline
    \end{tabular}
    }
    \caption{\small \it \textbf{Comparison between the new InstDet splits and Unified splits.} Eval inst$_{sc}$ and eval inst$_{uc}$ mean instances in the evaluation set with seen categories and unseen categories respectively. InstDet splits have more instances with unseen categories for better understanding the challenges in instance detection.}
    \label{tab:splits_comparison}
\end{table}

\begin{table}
% \vspace{-0.5cm}
\centering
\setlength{\tabcolsep}{2pt}
% \small
\resizebox{\columnwidth}{!}{
\begin{tabular}{c|c|cccc|cccc}
 &  & \multicolumn{4}{c|}{val} & \multicolumn{4}{c}{test} \\
backbone & method & AP & AP50 & AP50$_{se}$ & AP50$_{un}$ & AP & AP50 & AP50$_{se}$ & AP50$_{un}$ \\ \shline
\multirow{2}{*}{R50} & RPN+SFNet & 17.9 & 29.7 & 29.8 & 28.9 & 16.4 & 26.9 & 27.2 & 24.3 \\
 & \tadet & 18.4 & 35.0 & 35.3 & 31.3 & 19.6 & 37.1 & 37.4 & 33.6 \\ \hline
\multirow{2}{*}{R101} & RPN+SFNet & 18.5 & 30.6 & 30.9 & 27.6 & 16.9 & 27.4 & 27.7 & 24.4 \\
 & \tadet & 23.3 & 41.3 & 42.0 & 35.3 & 23.5 & 41.6 & 42.3 & 34.7 \\ \hline
\end{tabular}
}
% \vspace{-0.2cm}
\caption{\small \it \textbf{Instance-level detection benchmarking results on the InstDet splits of EgoObjects.} AP50$_{se}$ and AP50$_{un}$ are computed for instances with categories seen and unseen during training. The proposed \textup{\tadet} model significantly outperforms the baseline \textup{RPN+SFNet} approach.
% \fyx{Add some comments on our hypothesis}
}
\label{tab:inst_det}
% \vspace{-0.5cm}
\end{table}

% \section{Multiple samples for target registration}

\section{Discussions on CL benchmarks}
% \#1 multi-head architecture from tech report

% \#2 find a few papers exploring the architecture changes for CL
\subsection{CL instance detection.} 
The continual learning of the instance-level object detection is an under-explored area. 
The top submissions basically follow the similar strategy of applying continual learning techniques to a base category-level object detector and treating each instance as an individual category. 
For the base detector, both single-stage and two-stage detectors are applicable. The rank-1 team and the rank-3 team are using single-stage detectors (VarifocalNet \cite{varifocalnet} and FCOS \cite{tian-fcos2019}) while the rank-2 team is using two-stage detector (Faster R-CNN \cite{faster-rcnn}). And the detectors are pretrained on either COCO \cite{coco} or LVIS \cite{lvis} which are not the major factor of performance. For the network architecture, ResNet \cite{resnet} and its variants \cite{res2net} are still the dominating backbones. 
One key factor contributing to the performance is the replay buffer. 
All submissions observe that the use of full replay buffer is necessary for high performance. 
But they adopt different sampling strategies for filling the buffer. Specifically, rank-1 team samples an experience-balanced buffer, whereas rank-2 team and rank-3 team uses video-balanced buffer and instance-balanced buffer respectively. Distillation techniques are also widely used by rank-1 and rank-3 teams. 
However, these submissions still treat the instance detection as a close-set problem where the number of instances are fixed. This limits their adoption for real-world applications where the number of instances can scale up continuously. 

\subsection{CL category detection.} There are several interesting observations towards the top submissions. 
First, when dealing with the catastrophic forgetting, both the rank-1 and rank-3 teams adopted a uniform sampling strategy to cache data from previous tasks, with the difference in that the former approach samples history data in a video-wise manner while the latter did so on a frame basis. Unlike these two teams, the rank-2 team proposes to follow a ``Minority Replay'' strategy in which they sort and emphasis on selecting and buffering data from rare classes.  
In addition to replay buffer, the rank-1 team proposes to distill information from a teacher model trained in previous tasks to the model for the current task. 
However, these top submissions made limited changes to the detector architecture. The rank-1 team and rank-3 team directly adopt the VarifocalNet \cite{varifocalnet} and Faster R-CNN \cite{faster-rcnn}, respectively. The rank-2 team proposed to train a separate detection head for each experience, leading to a multi-head Faster R-CNN. This shows the potential of designing architectures tailored to the CL tasks. But the multi-head solution is not scalable. Therefore, there is still huge room for innovative ideas in this field. 
% As for the backbone network, most teams adopted the classical ResNet CNNs with the exception from the rank-3 team, where they ablated the comparisons between ResNets and Transformer architectures and chose Swin-Small \cite{liu-swin2021} as their backbone network. 

% \section{Video samples}
% We provide a group of 10 videos featuring the main object ``soccer ball'' in folder {\tt samples}.

\begin{figure}
    \centering
    \includegraphics[width=0.45\textwidth]{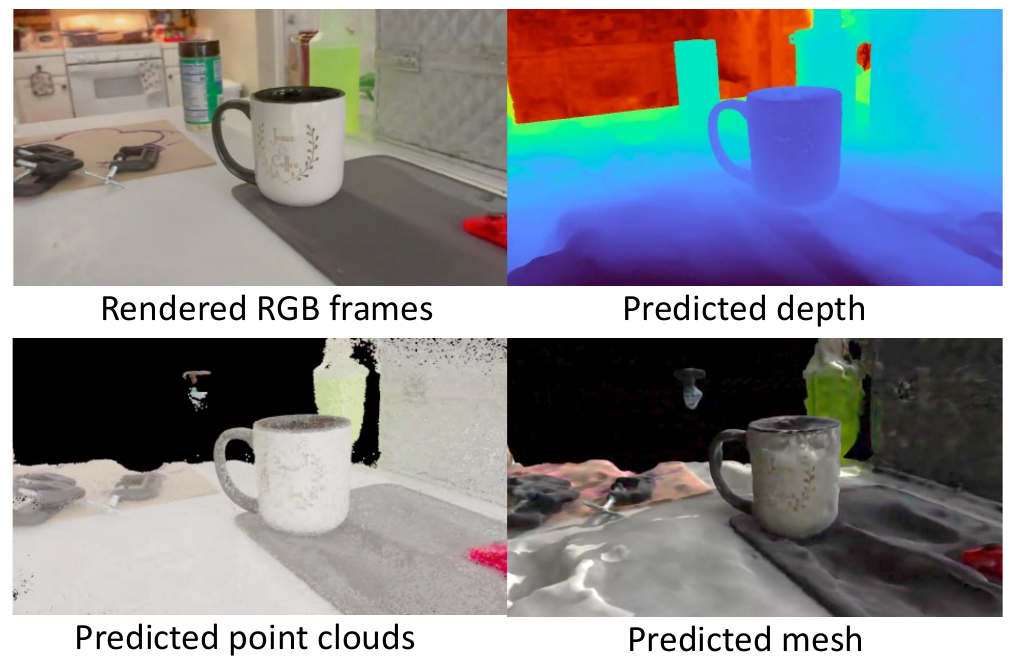}
    \caption{\small \it 
    \textbf{Predict 3D geometry on EgoObjects videos}. 
    We show sample frames from the video of {\tt 3d/coffeecup.mp4}. 
    }
    \label{fig:3d}
\end{figure}

\section{Predict 3D geometry on EgoObjects videos}
In this section, to demonstrate the potential of creating a 3D dataset from EgoObjects using a fully/semi-automatic pipeline,  
we leverage the latest development in neural scene representations to predict 3D geometry labels on EgoObjects videos~\cite{mildenhall-nerf2021,barron-mip2022}. 
We predict both the 3D point clouds and meshes on several example videos from EgoObjects.
% (see videos under {\tt /3d} directory).
To create the inputs for NeRF training, we sample video frames at 1 FPS and use NeRFacto~\cite{nerfacto} to train one model for each video.  
We render outputs at 15 FPS to generate all videos. 
Taking {\tt coffeecup.mp4} as an example, from left to right we present the rendered RGB frames, the estimated depth map, the predicted point clouds, and finally the predicted meshes (sample frames are presented in Figure~\ref{fig:3d}). 
Meshes are constructed from point clouds using Poisson surface reconstruction~\cite{kazhdan-poisson2006}.  
Given the 2D bounding box annotations and the predicted 3D point clouds and meshes, 
we can easily intersect the two to produce 3D geometry for specific objects,
which could potentially be used to train 3D bounding box detectors (\eg, CubeRCNN~\cite{brazil-omni3d2022}) and 3D mesh predictors (\eg, MeshRCNN~\cite{meshrcnn}). 

% \end{appendices}

\end{document}